\pdfoutput=1 
\documentclass[11pt]{article}

\usepackage[letterpaper, margin=1in]{geometry}
\usepackage[T1]{fontenc}
\usepackage[utf8]{inputenc}
\usepackage{lmodern}            
\usepackage{microtype}

\usepackage{amsmath, amssymb, amsthm}
\usepackage{mathtools}
\usepackage{bm}
\allowdisplaybreaks             

\usepackage{graphicx}
\graphicspath{{figures/}}       
\usepackage{booktabs}
\usepackage[font=small, labelfont=bf]{caption}
\usepackage{subcaption}
\usepackage{enumitem}
\usepackage{xcolor}

\usepackage[round]{natbib}

\usepackage[colorlinks=true, allcolors=linkblue]{hyperref}
\definecolor{linkblue}{RGB}{0, 51, 153}
\usepackage{algorithm}          
\usepackage{algorithmic}       
\usepackage[capitalize, noabbrev]{cleveref}   

\theoremstyle{plain}
\newtheorem{theorem}{Theorem}[section]
\newtheorem{lemma}[theorem]{Lemma}

\newtheorem{corollary}[theorem]{Corollary}

\theoremstyle{definition}

\newtheorem{assumption}{Assumption}           

\theoremstyle{remark}
\newtheorem{remark}[theorem]{Remark}

\crefname{assumption}{Assumption}{Assumptions}
\crefname{condition}{Condition}{Conditions}
\crefname{claim}{Claim}{Claims}

\numberwithin{equation}{section}              

\usepackage{authblk}

\newcommand{\keywords}[1]{\par\medskip\noindent\textbf{Keywords:} #1}

\title{Transfer Denoising of Low-Rank Plus Sparse Matrices \\ under Ambient and Rank Growth}

\author[$\sharp$]{
	Elynn Chen
}
\author[$\flat$]{
	Jinhang Chai
}
\author[$\diamond$]{
	Xuyuan Liu
}
\author[$\diamond$]{
	Yujun Yan \thanks{
		Corresponding authors: Y.~Yan:\texttt{Yujun.Yan@dartmouth.edu};
		E.~Chen: \texttt{elynn.chen@stern.nyu.edu}.
	}
}
\affil[$\sharp$]{Department of Technology, Operations, and Statistics, New York University}
\affil[$\flat$]{Department of Operations Research \& Financial Engineering, Princeton University}
\affil[$\diamond$]{Department of Computer Science, Dartmouth College}

\date{\today}

\begin{document}

\maketitle


\begin{abstract}
We study transfer learning for low-rank plus sparse matrix denoising when both ambient and latent dimensions grow. 
A data-rich source task provides subspace information for a higher-dimensional, data-scarce target with a small rank increment and localized sparse edits. 
We develop an anchored alternating-projection estimator that preserves transferred subspaces while learning target-specific coefficients, additional latent directions, and sparse modifications. 
We establish deterministic error bounds that separate target noise, representation growth, and source-estimation error, identifying improved rates through small structural increments and reduced noise within inherited subspaces. 
We further extend it to approximate anchoring quantifies robustness to subspace mismatch. 
We establish application-specific guarantees for Markov transition matrix estimation from single trajectories with dependent noise and for structured covariance estimation, with the latter analysis provided in the appendix. Experiments demonstrate consistent improvements over non-transfer baselines on synthetic data and an approximately 45\% reduction in principal-angle error on a NYC taxi transition-estimation task with expanding state spaces.

\keywords{Transfer learning; low-rank-plus-sparse matrix estimation; growing ambient dimensions; deterministic error bounds; .}
\end{abstract}

\section{Introduction}\label{sec:intro}

Modern learning systems rarely operate with fixed representations.
Feature and state spaces expand as new sensors, modalities, or modeling
components are introduced, while previously learned structure may remain
reusable. Such expansion creates a statistical asymmetry: historical
representations are learned from abundant source data, whereas the
enlarged target initially has limited samples. Retraining from scratch
discards information precisely when the increased dimensionality makes
estimation more demanding.

This paper studies \textbf{transfer learning for structured matrix
denoising under simultaneous ambient and representation growth}.
A known embedding maps source coordinates into a strictly larger target
space. We model the target through inherited source row and column
subspaces, additional low-dimensional directions, and sparse edits,
while allowing all target coefficients to change. The central question
is whether target estimation can depend on \emph{incremental innovation
complexity}, rather than the full expanded structure.

Classical multitask learning studies shared low-dimensional
representations in a common ambient space
\citep{ando2005framework,argyriou2008convex}, while domain adaptation
analyzes distribution shift under a fixed hypothesis class
\citep{ben2010theory}. Structured matrix estimation provides recovery
guarantees for single-task low-rank-plus-sparse models
\citep{candes2011robust,chandrasekaran2011rank,chen2021spectral}.
These results do not directly characterize the statistical benefit of
transferring estimated structure when both the ambient space and latent
representation expand.

We develop an \textbf{anchored constrained estimator} and a practical
alternating-projection procedure. Embedded source subspaces remain
fixed, while target coefficients, orthogonal innovations, and sparse
modifications are estimated from target data. Our deterministic analysis
separates inherited noise, representation growth, and source-estimation
error. The statistical guarantees concern the constrained estimator,
with an explicit optimization-error contribution for feasible approximate
minimizers. The deterministic result accommodates dependent matrix noise,
with observation-specific concentration yielding statistical rates for
covariance and Markov transition estimation.

\textbf{Novelty and technical challenges.}
We introduce a theoretical framework for
\emph{low-rank-plus-sparse matrix transfer under simultaneous ambient
and representation growth}, combining estimated source row and column
subspaces, freely re-estimated target coefficients, orthogonal
innovations, and sparse edits. The resulting deterministic theory
separates inherited-subspace noise, innovation complexity, sparse edits,
and source-anchor error, charging ambient-noise costs to
\emph{incremental rather than full target complexity}, while controlling
inherited coefficients through noise compressed to the transferred
subspaces.

Establishing this gain requires more than a generic rank-based bound,
which would discard the distinction between inherited and new
structure. Our analysis constructs a feasible source-anchored
comparison point, orthogonally decomposes the error relative to the
estimated anchors, and applies distinct noise controls to inherited
and innovation blocks. Quantitative low-rank--sparse separation
prevents cancellation between component errors, while source
estimation error enters as a separate perturbation rather than
inflated innovation complexity. These ingredients retain the
transfer-specific error structure even when source anchors are
estimated rather than known.

\textbf{Broad applicability and robustness.}
The motivating regime is one where \emph{ambient dimension grows faster
than intrinsic complexity}. Examples include persistent risk factors as
financial markets expand, shared biological factors across enlarged
genomic assays, and reusable transition modes as transportation networks
or reinforcement-learning state spaces grow. Our NYC taxi experiment
evaluates state-space-expansion transfer from 200 to 255 districts,
reducing subspace error by approximately 45\% relative to both
non-transfer baselines and supporting practical relevance beyond the
synthetic covariance setting.

The model accommodates rotations within inherited subspaces through
the freely re-estimated coefficient matrix $A^{(2)}$, interactions with
new directions through its cross blocks, and localized changes through
sparse edits. For aggregate anchoring tolerance $\epsilon$, the robust
extension of Theorem~\ref{thm:main}
(Appendix~\ref{appendix:robust}) adds a quadratic mismatch contribution
$O(\epsilon^2)\|A^{(2)}\|_2^2$, preserving the transfer advantage when
this contribution is controlled. Our rotation study
(Appendix~\ref{appendix:rotation}) complements this analysis:
substantial rotations can be tolerated when few directions change and
target samples are scarce, while principal-angle closeness alone does
not determine transferability.

\textbf{Contributions.}
$\bullet$ We formulate structured matrix transfer under
\emph{simultaneous ambient and representation growth}, with a constrained
estimator and a practical anchored alternating-projection procedure.

$\bullet$ We establish deterministic guarantees separating inherited
noise, innovation complexity, source-anchor error, and optimization
error, and quantify sensitivity to approximate anchoring.

$\bullet$ We specialize the framework to Markov transition estimation
from a single dependent trajectory and structured covariance estimation,
and evaluate transfer gains through synthetic, controlled-rotation,
and NYC taxi experiments.

\subsection{Related Work and Our Distinction}

\textbf{Multitask, Transfer, and Shared Representations.}
A large body of work studies multitask and transfer learning under the assumption that all tasks share a common representation and a fixed ambient parameter space \citep{ando2005framework,argyriou2008convex,ruder2017overview}. Domain adaptation theory similarly focuses on distribution shift under a fixed hypothesis class \citep{ben2010theory,ganin2016domain}, while multiple-source adaptation with limited labeled target data is studied by \citet{mansour2021theory}. These works do not analyze the growing two-sided subspace and sparse-edit structure considered here.
\textbf{In contrast}, our framework models asymmetric transfer with \emph{representation and ambient dimension growth}, where the source representation is embedded as a subspace of a larger target space. This structural asymmetry fundamentally distinguishes our setting from classical multitask and domain adaptation.

\textbf{Representation Learning and Feature Expansion.}
Representation learning aims to discover compact latent structure under fixed feature spaces \citep{bengio2013representation}. Continual and lifelong learning study how models evolve as new tasks arrive \citep{kirkpatrick2017overcoming,lopez2017gradient,chen2018lifelong}, and architectural approaches such as progressive networks explicitly expand model capacity over time \citep{rusu2016progressive}. Theoretical analyses of shared representation learning characterize benefits across tasks and the role of task diversity \citep{maurer2016benefit,tripuraneni2020theory}.
Our work studies \emph{representation expansion} for structured matrices, quantifying how reusable source subspaces reduce target estimation complexity when new latent directions and sparse changes remain limited.

\textbf{Structured Matrix Estimation.}
Low-rank and low-rank-plus-sparse matrix estimation provides recovery guarantees under fixed ambient dimensions \citep{candes2011robust,chandrasekaran2011rank,agarwal2012noisy,chen2021bridging,chen2021spectral}, and recent work allows arbitrary or dependent entrywise noise \citep{agterberg2022entrywise,chai2025structured}.
Prior-subspace information is exploited in robust PCA \citep{zhan2015robust}, and matrix transfer with missing entire rows or columns is studied by \citet{jalan2025optimal}. These are relevant precedents, but their recovery and observation models differ from our noisy, two-sided subspace-anchored denoising problem with sparse edits. Our analysis tracks inherited-block noise, orthogonal innovations, and source-estimation errors jointly.

\textbf{Applications to Reinforcement Learning and Structured Dynamics.}
Structured representations are widely used in reinforcement learning, particularly for low-rank or compressed Markov transitions \citep{zhang2019spectral}. Transfer Q-learning with composite MDP structures shares low-rank transition structure while allowing sparse task differences \citep{chai2025transition}. Our Markov application instead studies estimation of an enlarged frequency matrix, with both subspace innovations and sparse edits under a known state embedding.

\section{Problem Setup and Transfer Model}\label{sec:model}

We consider a sequence of related learning tasks indexed by $m = 1, \dots, M$, where task $M$ is the target of interest and tasks $1, \dots, M-1$ serve as source tasks.
 For each task $m$, we observe a noisy matrix
\begin{equation}\label{eqn:denoising}
Y^{(m)} = \Theta^{(m)} + W^{(m)}, \qquad \Theta^{(m)} \in \mathbb{R}^{p_m \times q_m},
\end{equation}
where $\Theta^{(m)}$ denotes the unknown task-specific parameter matrix and $W^{(m)}$ is an additive noise matrix. The ambient dimensions $(p_m, q_m)$ are allowed to vary across tasks. Our goal is to estimate the target parameter $\Theta^{(M)}$ by leveraging information from previously observed tasks $\{Y^{(m)}\}_{m<M}$, in addition to the target observation $Y^{(M)}$. We focus on transfer settings in which the tasks are related through their underlying structure, so that knowledge learned from earlier tasks can reduce the sample or estimation complexity of the target task.

Although the framework applies to general $M$, for clarity of exposition and theoretical analysis we specialize to $M = 2$ in the remainder of the paper; extensions to multiple sequential tasks follow naturally.

\subsection{Ambient Feature Space Growth}

A central feature of our setting is that the ambient feature space may expand across tasks. We focus on structured expansion scenarios in which existing features are preserved and embedded within a larger ambient space.

Formally, the source task parameter satisfies
$
\Theta^{(1)} \in \mathbb{R}^{p_1 \times q_1},
$
while the target task parameter satisfies
$
\Theta^{(2)} \in \mathbb{R}^{p_2 \times q_2},
\; p_2 \ge p_1,\; q_2 \ge q_1.
$
To compare objects across dimensions, we introduce an \emph{embedding operator}:
\[
B:\mathbb R^{p_1\times q_1}\to\mathbb R^{p_2\times q_2}, \quad B(A)=\begin{bmatrix}A&0\\0&0\end{bmatrix},
\]
which embeds a source matrix into the target ambient space by padding zeros outside a designated $p_1 \times q_1$ block. This model captures ambient feature expansion in which new coordinates are appended rather than replacing existing ones, motivating transfer methods that explicitly reuse structure learned in the lower-dimensional source ambient space.

\subsection{Representation Growth: Low-Rank and Sparse Transfer Structure}\label{sec:representation_growth}

Within the expanded ambient feature space, we model how the {\it intrinsic representation} evolves across tasks. We assume that each task parameter admits a structured decomposition
$$
\Theta^{(m)} = L^{(m)} + S^{(m)},
$$
where $L^{(m)}$ captures low-dimensional latent structure and $S^{(m)}$ captures sparse, task-specific deviations.

\textbf{Low-rank representation growth (anchored subspace expansion).}
Let the source low-rank component admit the population factorization
$$
L^{(1)} = U^{(1)} \Sigma^{(1)} (V^{(1)})^\top,\;
U^{(1)} \in \mathbb{R}^{p_1 \times r_1},\;
V^{(1)} \in \mathbb{R}^{q_1 \times r_1}.
$$
After embedding into the target ambient space, the source row and column subspaces become
$$
B(U^{(1)}) \subset \mathbb{R}^{p_2\times r_1}, \qquad B(V^{(1)}) \subset \mathbb{R}^{q_2\times r_1}.\footnote{We overload $B(\cdot)$ to denote zero-padding of matrices or factor matrices along the appropriate dimensions, depending on context.}
$$
We model the target low-rank component by {\it expanding these embedded subspaces with a small number of new directions}. Specifically, we assume that the column and row spaces of $L^{(2)}$ can be written as
$$
U^{(2)} = [\, B(U^{(1)}) \;\; U_\Delta \,], \qquad
V^{(2)} = [\, B(V^{(1)}) \;\; V_\Delta \,],
$$
where $U_\Delta \in \mathbb{R}^{p_2 \times \delta_{r,2}}$ and $V_\Delta \in \mathbb{R}^{q_2 \times \delta_{r,2}}$ have orthonormal columns and satisfy
$
U_\Delta^\top B(U^{(1)}) = 0$ and $V_\Delta^\top B(V^{(1)}) = 0.
$
The target low-rank component is then parameterized as $L^{(2)} = U^{(2)} A^{(2)} (V^{(2)})^\top$
for some coefficient matrix $A^{(2)} \in \mathbb{R}^{r_2 \times r_2}$, where $r_2 = r_1 + \delta_{r,2}$, $0<\delta_{r,2}<r_1$ and $\delta_{r,2}\ll \delta_{p,2}$.

\textbf{Inherited-innovation interactions.}
Set $U_0=B(U^{(1)})$ and $V_0=B(V^{(1)})$, and partition $A^{(2)}$ into blocks $A_{ij}$. Then
\[
\begin{aligned}
L^{(2)}={}&U_0A_{11}V_0^\top+U_0A_{12}V_\Delta^\top
+U_\Delta A_{21}V_0^\top+U_\Delta A_{22}V_\Delta^\top.
\end{aligned}
\]
The off-diagonal blocks allow interactions between inherited and new directions. All coefficients are re-estimated; neither block diagonality nor preservation of source singular values is assumed.

Under this construction, the embedded source subspaces are {\it explicitly preserved}, and any additional latent structure appears only through the orthogonal innovation subspaces spanned by $U_\Delta$ and $V_\Delta$. We therefore define {\it representation growth} as this low-dimensional expansion of the row and column subspaces of the low-rank component relative to the embedded source representation, rather than merely an increase in matrix rank. The factor increment $\delta_{r,2}$ is assumed small compared to the ambient feature expansion.

\begin{remark}[Identifiability]
Given the embedding $B(\cdot)$, decomposing the target low-rank component into an embedded source subspace and an orthogonal innovation subspace is without loss of generality, up to standard low-rank indeterminacies. Without such an embedding, the source subspace would be unidentifiable in the enlarged ambient space.
\end{remark}

\textbf{Sparse structure transfer.}
We allow the sparse component to evolve via localized changes:
$$
S^{(2)} = B\!\left(S^{(1)}\right) + \Delta^{(2)}_S,
\qquad \|\Delta^{(2)}_S\|_0 \le \delta_{s,2},
$$
where $\delta_{s,2} \ll \|S^{(1)}\|_0$, modeling sparse edits that affect only a small number of entries in the expanded ambient space.

Together, these assumptions describe a transfer regime in which the target task inherits most of its structure from the source task, with only low-dimensional representation innovations and sparse modifications introduced by the expanded feature set. This structure underlies both the anchored estimator in Section~\ref{sec:estimation} and the improved statistical rates established in Section~\ref{sec:theory}.

\section{Estimation Method: Anchored Low-Rank Plus Sparse Transfer} \label{sec:estimation}

We assume the source task is estimated from abundant source data, yielding estimates $\widehat L^{(1)}$ and $\widehat S^{(1)}$ of the low-rank and sparse components. Let
$
\widehat L^{(1)} = \widehat U^{(1)} \widehat \Sigma^{(1)} (\widehat V^{(1)})^\top
$
denote a rank-$r_1$ factorization of the source low-rank estimate, and define the embedded source subspaces in the target ambient space as
$$
\widetilde U^{(1)} := B(\widehat U^{(1)}) \in \mathbb{R}^{p_2 \times r_1},
\;
\widetilde V^{(1)} := B(\widehat V^{(1)}) \in \mathbb{R}^{q_2 \times r_1},
$$
together with the embedded sparse anchor $S_0 := B(\widehat S^{(1)})$.

Given $Y^{(2)}$, we fit a low-rank component represented in the estimated source spans plus at most $\delta_{r,2}$ innovation directions on each side, and a sparse component differing from $S_0$ in at most $\delta_{s,2}$ entries. These are constraints on the fitted model; population departures induced by source-estimation error are accounted for in the analysis.

Formally, given inputs $Y^{(2)}$, target hyperparameters $\delta_{r,2}$ and $\delta_{s,2}$, and incoherence bound $\mu$ defining the statistical feasible set, consider
\begin{gather} \label{eqn:optimization-problem}
\min_{U_\Delta,V_\Delta,A^{(2)},S}\
\frac12\left\|Y^{(2)}-\Big(U^{(2)}A^{(2)}(V^{(2)})^\top + S\Big)\right\|_F^2 \\
\quad \text{s.t.} \ \ U^{(2)}=[\widetilde U^{(1)}\ U_\Delta],\quad
V^{(2)}=[\widetilde V^{(1)}\ V_\Delta], \nonumber \\
\qquad (U^{(2)})^\top U^{(2)}=(V^{(2)})^\top V^{(2)}=I_{r_2}, \nonumber \\
\qquad \|S-S_0\|_0\le \delta_{s,2}, \quad U^{(2)}A^{(2)}(V^{(2)})^\top \in \mathcal{M}_{\mu}. \nonumber
\end{gather}
This formulation makes explicit that only the {\it innovation subspaces} $(U_\Delta, V_\Delta)$, the associated coefficients, and the {\it sparse edits} relative to $S_0$ are estimated from the target data. The embedded source structure is treated as fixed, serving as an anchor that reduces the effective complexity of the target.

\textbf{Statistical estimator and practical solver.}
Let $\mathcal F$ be the feasible set in \eqref{eqn:optimization-problem} and $\mathcal L(L,S)=\tfrac12\|Y^{(2)}-L-S\|_F^2$. Our statistical estimator is a global minimizer over $\mathcal F$, or a feasible approximate minimizer satisfying
\begin{equation}\label{eq:approx-optimality}
\mathcal L(\widehat L^{(2)},\widehat S^{(2)})
\le \inf_{(L,S)\in\mathcal F}\mathcal L(L,S)+\varepsilon_{\mathrm{opt}}.
\end{equation}
The basic-inequality analysis retains an additive optimization-error
contribution. Algorithm~\ref{alg:transfer-altproj} is the practical
procedure used in our experiments; its stopping criterion does not
certify \eqref{eq:approx-optimality} or enforcement of the incoherence
constraint. This distinction between estimator-level statistical
guarantees and solver-level convergence guarantees follows established
practice in nonconvex learning and structured matrix estimation
\citep{jain2017nonconvex,chai2025structured}.

\subsection{Anchored Low-Rank and Sparse Projections}

For computation, we alternate an anchored spectral update, which retains the embedded source spans while estimating innovation directions, with the exact sparse-edit projection. The spectral update is not claimed to be the global solution of the constrained low-rank subproblem.

\textbf{Anchored low-rank update.} Given $M\in\mathbb R^{p_2\times q_2}$, write $\mathsf P_{\text{anch-rank}}(M;\widetilde U^{(1)},\widetilde V^{(1)},\delta_{r,2},r_1)$ for the following spectral update. It constructs an anchored rank-at-most-$r_2$ approximation. With unrestricted inherited--innovation interactions, selecting directions from the doubly projected residual alone need not minimize the full constrained Frobenius loss.

The update selects innovation directions orthogonal to the estimated source spans and subsequently refits all coefficient blocks. Define the orthogonal projectors
\begin{equation}\label{eqn:anchored_projection}
P_{\widetilde U} := \widetilde U^{(1)} (\widetilde U^{(1)})^\top,
\qquad
P_{\widetilde V} := \widetilde V^{(1)} (\widetilde V^{(1)})^\top.
\end{equation}
The innovation component is obtained by projecting $M$ onto the orthogonal complement,
$
(I - P_{\widetilde U})\, M\, (I - P_{\widetilde V}),
$
and extracting its leading $\delta_{r,2}$ singular left and right directions, denoted $U_\Delta$ and $V_\Delta$.
The updated factors $\widehat U^{(2)} = [\,\widetilde U^{(1)}\ \ U_\Delta\,]$ and $\widehat V^{(2)} = [\,\widetilde V^{(1)}\ \ V_\Delta\,]$ explicitly enforce the anchoring constraint. The coefficient matrix is then $\widehat A^{(2)} = (\widehat U^{(2)})^\top M \widehat V^{(2)}$, and the update returns
$\widehat L^{(2)} := \widehat U^{(2)} \widehat A^{(2)} (\widehat V^{(2)})^\top$. The full procedure is in Algorithm \ref{alg:anchored-lowrank-proj} (Appendix \ref{append:algo}).

\textbf{Sparse-edit projection.}
Given $M \in \mathbb{R}^{p_2 \times q_2}$, we define the sparse-edit projection relative to the transferred anchor $S_0$ as
$
\mathsf{P}_{\mathrm{edit}}(M; S_0, \delta_{s,2})
:= S_0 + \mathcal{H}_{\delta_{s,2}}(M - S_0),
$
where $\mathcal{H}_k(\cdot)$ retains the $k$ largest-magnitude entries and sets the rest to zero. This operator projects $M$ onto the set of matrices that differ from $S_0$ in at most $\delta_{s,2}$ entries, corresponding to sparse edits relative to the source task. See Algorithm \ref{alg:sparse-edit-proj} (Appendix \ref{append:algo}).

Together, these two updates target the {\it low-rank subspace innovations} and {\it sparse modifications} introduced by the expanded feature space, while anchoring the estimation to structure learned from the source task.

\subsection{Transfer Alternating Projection Algorithm}

Starting from the transferred sparse anchor $S_0 = B(\widehat S^{(1)})$, our algorithm alternates between two updates. The {\bf sparse-edit update} sets
$
S_{t+1} = \mathsf{P}_{\mathrm{edit}}\bigl(Y^{(2)} - L_t;\, S_0, \delta_{s,2}\bigr),
$
estimating only sparse deviations relative to $S_0$. The {\bf anchored low-rank update} forms the residual $M_{t+1} = Y^{(2)} - S_{t+1}$ and applies
$
L_{t+1} = \mathsf P_{\text{anch-rank}}(M_{t+1};\widetilde U^{(1)},\widetilde V^{(1)},\delta_{r,2},r_1).
$
The complete procedure is in Algorithm~\ref{alg:transfer-altproj}; hats on its outputs denote numerical estimates, which are not automatically the constrained statistical estimator.

\begin{algorithm}[h!]
\caption{Transfer AltProj for Matrix Estimation}
\label{alg:transfer-altproj}
\begin{algorithmic}[1]
\REQUIRE $Y^{(2)}\in\mathbb{R}^{p_2\times q_2}$; anchored subspaces
$\widetilde U^{(1)}\in\mathbb{R}^{p_2\times r_1}$,
$\widetilde V^{(1)}\in\mathbb{R}^{q_2\times r_1}$; anchor $S_0=B(\widehat S^{(1)})$;
rank increment $\delta_{r,2}$; sparse edit budget $\delta_{s,2}$; tolerance $\varepsilon$.
\ENSURE $\widehat L^{(2)}$, $\widehat S^{(2)}$.

\STATE $r_2 \leftarrow r_1+\delta_{r,2}$;\
$P_{\widetilde U} \leftarrow \widetilde U^{(1)}\widetilde U^{(1)\top}$,\
$P_{\widetilde V} \leftarrow \widetilde V^{(1)}\widetilde V^{(1)\top}$;\ $L_0 \leftarrow 0$.

\FOR{$t=0,1,2,\dots$}
\STATE $S_{t+1} \leftarrow \mathsf{P}_{\mathrm{edit}}\bigl(Y^{(2)} - L_t;\, S_0, \delta_{s,2}\bigr)$;\
$M_{t+1} \leftarrow Y^{(2)} - S_{t+1}$.
\STATE $L_{t+1} \leftarrow \mathsf P_{\text{anch-rank}}(M_{t+1};\widetilde U^{(1)},\widetilde V^{(1)},\delta_{r,2},r_1)$.

\STATE \textbf{break} if
$\frac{\|L_{t+1}-L_t\|_F}{\max\{\|L_t\|_F,10^{-12}\}}
+\frac{\|S_{t+1}-S_t\|_F}{\max\{\|S_t\|_F,10^{-12}\}}
\le \varepsilon$.
\ENDFOR
\STATE \textbf{return} $\widehat L^{(2)}\leftarrow L_{t+1}$, \ $\widehat S^{(2)}\leftarrow S_{t+1}$.
\end{algorithmic}
\end{algorithm}

\subsection{Computational Complexity and Practical Considerations}

The sparse-edit step can select the $\delta_{s,2}$ largest-magnitude entries in $O(p_2q_2)$ time; even for a small edit budget, a dense residual must be scanned. The innovation step requires only $\delta_{r,2}$ leading singular directions, and can use Lanczos or randomized SVD. For a dense residual, a fixed-pass randomized implementation with sketch width $\ell\ge\delta_{r,2}$ costs
\[
O\!\left(p_2q_2\ell+(p_2+q_2)(r_1\ell+\ell^2)\right),
\]
including implicit application of the source projectors and orthogonalization \citep{martinsson2020randomized}. Lanczos costs additionally depend on accuracy and spectral separation. Coefficient refitting and dense reconstruction contribute up to $O(p_2q_2r_2+(p_2+q_2)r_2^2)$ per iteration. Thus, the singular-vector search is reduced from target rank to innovation rank, but total runtime also includes anchor, matrix-product, and sparse-update costs; a uniform speedup is not guaranteed.

The source anchors are computed once. The practical procedure does not take $\mu$ as an input or explicitly enforce the incoherence constraint used to define the statistical estimator. Rank and edit budgets can be selected by held-out validation or domain knowledge; spectral diagnostics can guide the search but require signal separation for reliable rank identification. We do not establish global convergence or a computable objective-gap certificate for Algorithm~\ref{alg:transfer-altproj}.

\section{Theoretical Guarantees}\label{sec:theory}

We analyze feasible approximate minimizers of
\eqref{eqn:optimization-problem} satisfying \eqref{eq:approx-optimality}.
The main bound is deterministic; probability enters only through
observation-specific noise bounds in the applications.

\textbf{Notation and regularity.}
We use $\|\cdot\|_F$, $\|\cdot\|_2$, and $\|\cdot\|_{\max}$ for
Frobenius, spectral, and entrywise maximum norms;
$\|M\|_{1,1}=\sum_{i,j}|M_{ij}|$ and
$\|M\|_{2,\infty}=\max_i\|M_{i,:}\|_2$.
The notation $\lesssim$ hides universal constants unless otherwise stated.
For the rank budget $r_2$, $\mathcal M_\mu$ consists of matrices admitting
$r_2$-column orthonormal singular-vector frames $Q_L,Q_R$ with
\[
\|Q_L\|_{2,\infty}\le\mu\sqrt{r_2/p_2},\qquad
\|Q_R\|_{2,\infty}\le\mu\sqrt{r_2/q_2}.
\]
Zero singular values are allowed. Set $U_0=B(U^{(1)})$ and
$V_0=B(V^{(1)})$. Let $\bar P_U,\bar P_V\in\mathcal O_{r_1,r_1}$
be Frobenius-optimal orthogonal alignments of $U_0,V_0$ to the
estimated anchors. Define
\[
D_U=U_0\bar P_U-\widetilde U^{(1)},\qquad
D_V=V_0\bar P_V-\widetilde V^{(1)},
\]
$\alpha_U=\|D_U\|_F$, $\alpha_V=\|D_V\|_F$,
$e_U=\|D_U\|_2$, and $e_V=\|D_V\|_2$.
These population alignments are proof devices, not algorithmic inputs.

\begin{assumption}[Structural and source-anchor regularity]
\label{asp:incoherence-sparsity}
The exact transfer model of Section~\ref{sec:representation_growth}
holds with $\operatorname{rank}(L^{(2)})=r_2=r_1+\delta_{r,2}$.
For some $\mu_0>0$, use the fitting budget $\mu=2\mu_0$.

\textbf{(i) Row regularity.} The target factor frames and embedded
source estimates have orthonormal columns and satisfy
\[
\begin{aligned}
\max\{\|U^{(2)}\|_{2,\infty},\|\widetilde U^{(1)}\|_{2,\infty}\}
&\le\mu_0\sqrt{r_2/p_2},\\
\max\{\|V^{(2)}\|_{2,\infty},\|\widetilde V^{(1)}\|_{2,\infty}\}
&\le\mu_0\sqrt{r_2/q_2}.
\end{aligned}
\]

\textbf{(ii) Source accuracy.} $\max\{e_U,e_V\}\le1/2$.

\textbf{(iii) Sparse edits.} For a sufficiently small universal $c>0$,
\[
\|S^{(2)}-B(S^{(1)})\|_0\le\delta_{s,2}
\le\frac{c\min\{p_2,q_2\}}{\mu^2r_2^3}.
\]
\end{assumption}

These are sufficient conditions: estimated-anchor regularity is not
implied by population incoherence alone. Lemma~\ref{lem:comparison-regularity}
constructs a feasible source-adjusted comparison pair.
Lemma~\ref{lem:rectangular-separation} derives from
\citet{chai2025structured} the strict separation
\begin{equation}\label{eq:transfer-separation}
\|D+H\|_F^2\ge\tfrac12(\|D\|_F^2+\|H\|_F^2),
\end{equation}
for $D=L-L'$ with $L,L'\in\mathcal M_\mu$ and
$\|H\|_0\le2\delta_{s,2}$.
Our convention places $\mu$ outside the square root, hence the
$\mu^2$ in (iii). These conservative conditions are not necessary for
empirical success.

\textbf{Main deterministic bound.}
Write
\begin{align*}
\mathcal E_{\mathrm{tgt}}={}&
r_1\|\widetilde U^{(1)\top}W^{(2)}\widetilde V^{(1)}\|_2^2
+\delta_{r,2}\|W^{(2)}\|_2^2
+\delta_{s,2}\|W^{(2)}\|_{\max}^2,\\
\mathcal E_{\mathrm{src}}={}&
(\alpha_U^2+\alpha_V^2)\|A^{(2)}\|_2^2
+\|\widehat S^{(1)}-S^{(1)}\|_F^2.
\end{align*}
\begin{theorem}[Deterministic transfer bound]\label{thm:main}
Under Assumption~\ref{asp:incoherence-sparsity}, every feasible solution
satisfying \eqref{eq:approx-optimality} obeys
\begin{equation}\label{eq:main-transfer-bound}
\|\widehat L^{(2)}-L^{(2)}\|_F^2+\|\widehat S^{(2)}-S^{(2)}\|_F^2\\
\; \lesssim \; \mathcal E_{\mathrm{tgt}}+
\mathcal E_{\mathrm{src}}+\varepsilon_{\mathrm{opt}}.
\end{equation}
An exact constrained minimizer has $\varepsilon_{\mathrm{opt}}=0$.
\end{theorem}

The proof in Appendix~\ref{appendix:proof_main} decomposes the matrix
error using the \emph{estimated} anchor projectors, retaining exact
block orthogonality despite source error. The inherited block pays for
compressed noise; the other blocks pay ambient spectral-noise costs
at ranks $O(\delta_{r,2})$. Source and optimization errors remain
separate. The theorem does not certify Algorithm~\ref{alg:transfer-altproj}.

\begin{corollary}[Source-error specialization]\label{thm:chai2025_(1)}
Under Theorem~\ref{thm:main}'s assumptions, let $\sigma_{r_1}>0$ be
the smallest nonzero singular value of $L^{(1)}$. Suppose
\[
\begin{aligned}
&\|\widehat L^{(1)}-L^{(1)}\|_F^2+
\|\widehat S^{(1)}-S^{(1)}\|_F^2\lesssim R_1,\\
&R_1=r_1\|W^{(1)}\|_2^2+s_1\|W^{(1)}\|_{\max}^2,
\end{aligned}
\]
where $s_1=\|S^{(1)}\|_0$, and
$\|\widehat L^{(1)}-L^{(1)}\|_2\le\sigma_{r_1}/2$.
Then the right-hand side of \eqref{eq:main-transfer-bound} is at most
\[
C\left\{\mathcal E_{\mathrm{tgt}}+
R_1\left(1+\frac{\|A^{(2)}\|_2^2}{\sigma_{r_1}^2}\right)
+\varepsilon_{\mathrm{opt}}\right\}.
\]
\end{corollary}
The corollary follows from Wedin's singular-subspace bound
\citep{chen2021spectral}; the row-regularity assumptions remain required.
Thus source error is lower order only when its displayed contribution
is small relative to target statistical error.

\textbf{Approximate anchoring.}
Theorem~\ref{thm:main-robust} in Appendix~\ref{appendix:robust} replaces
exact inheritance by population anchor discrepancies
$\epsilon_U,\epsilon_V$. Under the stated row-regularity and combined
proximity conditions, it adds
$(\epsilon_U^2+\epsilon_V^2)\|A^{(2)}\|_2^2$ to
\eqref{eq:main-transfer-bound}. This is an explicit approximation cost,
not an automatically negligible term. Small principal angles alone
do not imply low innovation rank.

\textbf{Quantitative comparison.}
For target-only constrained estimation, under the corresponding
\emph{full-sparsity} regularity condition, the benchmark is
\citep{chai2025structured}
\[
r_2\|W^{(2)}\|_2^2+s_2\|W^{(2)}\|_{\max}^2,
\qquad s_2=\|S^{(2)}\|_0\le s_1+\delta_{s,2}.
\]
Equality in the sparsity relation need not hold: edits can modify or
remove existing entries. For square Gaussian matrix noise with
independent entries of variance $v_2$, independent of the source data,
the respective leading conditional expected-error bounds are
\[
\begin{aligned}
\text{transfer: }&v_2\{r_1^2+\delta_{r,2}p_2+
\delta_{s,2}\log(e p_2)\},\\
\text{target only: }&v_2\{r_2p_2+s_2\log(e p_2)\}.
\end{aligned}
\]
Transfer additionally pays $\mathcal E_{\mathrm{src}}+
\varepsilon_{\mathrm{opt}}$. The inherited-noise reduction is of order
$p_2/r_1$ in this isotropic example. These are comparisons of
\emph{error-bound scalings}, not pointwise risk orderings or a minimax
claim. Markov and covariance observations require the separate
concentration arguments below and in Appendix~\ref{appendix:pca_application}.

\section{Application to Markov Transition Matrix Estimation}
\label{sec:transition-matrix-estimation}

We apply our subspace-anchored transfer framework to Markov transition matrix estimation from a single trajectory with expanding state spaces, casting the problem as denoising an empirical frequency matrix with dependent noise. The key technical challenge is that the ``noise'' induced by a single Markov path is \emph{entrywise dependent}, so classical i.i.d.\ concentration does not apply.

Let the source chain evolve on state space $[p_1]$ and the target chain on $[p_2]$, where $p_2=p_1+\delta_{p,2}$. For each task $m\in\{1,2\}$, let $P^{(m)}$ denote the transition matrix, $\pi^{(m)}$ the stationary distribution, and
$
F^{(m)} := \operatorname{Diag}(\pi^{(m)})P^{(m)}
$
the corresponding frequency matrix. We observe a single stationary trajectory of length $n_m$ from each chain and form the empirical frequency matrices
$
\widehat F^{(m)} := \frac{1}{n_m}\sum_{t=0}^{n_m-1} e_{X^{(m)}_t}e_{X^{(m)}_{t+1}}^\top.
$
By construction,
\begin{equation}\label{eqn:markov-denoising}
\widehat F^{(m)} = F^{(m)} + W^{(m)},
\end{equation}
where $W^{(m)}$ denotes the estimation noise induced by finite sampling. Since all entries of $\widehat F^{(m)}$ are computed from a single Markov trajectory, the noise matrix $W^{(m)}$ exhibits strong entrywise dependence, and classical i.i.d.\ concentration arguments do not apply. Nevertheless, this representation places Markov transition estimation exactly in the denoising setting of Section~\ref{sec:model}, allowing deterministic guarantees to be applied once suitable bounds on $\|W^{(m)}\|$ and $\|W^{(m)}\|_{\max}$ are available.

We assume that the frequency matrices admit low-rank plus sparse decompositions $F^{(m)}=L^{(m)}+S^{(m)}$, and that $(F^{(1)},F^{(2)})$ satisfy the transfer structure of Section~\ref{sec:model}: the target frequency matrix inherits the embedded source structure via $B$, with low-rank subspace growth of at most $\delta_{r,2}$ new directions and sparse edits affecting at most $\delta_{s,2}$ entries.

\textbf{Estimator: transfer denoising of $\widehat F$ and plug-in $\widehat P$.}
Given the empirical frequency matrices $\widehat F^{(m)}$ for $m=1,2$, we estimate $F^{(2)}$ by treating $\widehat F^{(2)}$ as the observation matrix in the denoising model and invoking Algorithm~\ref{alg:transfer-altproj}, using the anchored subspaces and sparse anchor learned from the source task. Algorithm~\ref{alg:transfer-altproj} yields
$
\widehat F_{\mathrm{tran}}^{(2)} = \widehat L^{(2)} + \widehat S^{(2)}.
$
We then employ a standard plug-in estimator: estimate the stationary distribution by $\widehat \pi^{(2)} = \widehat F_{\mathrm{tran}}^{(2)}\mathbf{1}$ and define
\begin{equation}\label{eq:P-hat}
\widehat P_{\mathrm{tran}}^{(2)} = \operatorname{Diag}(\widehat \pi^{(2)})^{-1}\widehat F_{\mathrm{tran}}^{(2)},
\end{equation}
projecting onto the probability simplex row-wise when necessary to ensure stochasticity. The resulting estimator leverages structural information transferred from the source chain while adapting to the expanded state space using only target-task data.

\textbf{Theoretical guarantee.}
Combining Theorem~\ref{thm:main} with recent concentration results for Markov frequency matrices from a single trajectory yields an end-to-end statistical guarantee.

\begin{corollary}[Transfer Markov transition estimation]
\label{cor:transfer-markov}
Let $\tau_\star$ denote the $(1/4)$-mixing time of the target chain and define $\pi^{(2)}_{\min}=\min_i \pi^{(2)}_i$ and $\pi^{(2)}_{\max}=\max_i \pi^{(2)}_i$. Under Assumption~\ref{asp:markov-structured} in Appendix~\ref{appendix:markov_application}, there exist constants $c,c_0>0$ such that, with probability at least $1-n_2^{-c_0}$, the output $(\widehat L^{(2)},\widehat S^{(2)})$ of Algorithm~\ref{alg:transfer-altproj} satisfies
\begin{equation*}
\begin{aligned}
& \|\widehat F^{(2)}_{\emph{tran}} - F^{(2)}\|_F^2 \\
& \lesssim \frac{\pi_{\max}^{(2)}\tau_\star \log^2 n_2}{n_2}
\Big(r_1^2 + \delta_{r,2}p_2 + \delta_{s,2}\log p_2\Big) 
+
\mathcal{E}_{\emph{src}},
\end{aligned}
\end{equation*}
where $\mathcal{E}_{\mathrm{src}}$ collects the source-anchor error terms from Theorem~\ref{thm:main}.

Moreover, if the estimated stationary distribution $\widehat\pi^{(2)}=\widehat F^{(2)}_{\mathrm{tran}}\mathbf{1}$ satisfies $\min_i \widehat\pi^{(2)}_i\ge \pi^{(2)}_{\min}/2$, then the plug-in transition estimator \eqref{eq:P-hat} obeys
\begin{align*}
\|\widehat P^{(2)}_{\mathrm{tran}} - P^{(2)}_{\mathrm{tran}}\|_F^2
\;\lesssim\;
(\pi^{(2)}_{\min})^{-2}\,\|\widehat F^{(2)}_{\mathrm{tran}} - F^{(2)}\|_F^2, \\
\;\;
\|\widehat P^{(2)}_{\mathrm{tran}} - P^{(2)}_{\mathrm{tran}}\|_{1}
\;\lesssim\;
(\pi^{(2)}_{\min})^{-1}\,\|\widehat F^{(2)}_{\mathrm{tran}} - F^{(2)}\|_{1}.
\end{align*}
\end{corollary}

\begin{remark}[Comparison to non-transfer and regime interpretation]
\label{rem:markov-interpret}
Corollary~\ref{cor:transfer-markov} shows that transfer replaces the ambient dependence on $(r_2,s_2)$ by the incremental complexities $(\delta_{r,2},\delta_{s,2})$, plus an anchored subspace term $r_1^2$ from the inherited ``old-old'' block; when $p_2\gg r_1$, this term is typically dominated. Direct estimation of $F^{(2)}$ at the same sample size $n_2$ incurs the larger leading term
$
\frac{\pi_{\max}^{(2)}\tau_\star \log^2 n_2}{n_2}\big(r_2 p_2 + s_2\log p_2\big),
$
which is strictly worse whenever $r_2\gg \delta_{r,2}$ and $s_2\gg \delta_{s,2}$.
\end{remark}

\section{Empirical Validation}
\label{sec:empirical_application}

We validate the framework on two studies: a synthetic structured-covariance task that isolates the role of incremental rank, and a real-data Markov transition task on NYC taxi trips that tests the framework under genuine ambient growth.

\subsection{Synthetic Covariance Estimation under Enlarged Dimension}
\label{sec:empirical_synthetic}

For each task $m\in\{1,2\}$, we observe i.i.d.\ samples $X^{(m)}_1,\dots,X^{(m)}_{n_m}\in\mathbb R^{p_m}$ with mean zero and covariance $\Sigma^{(m)} = L^{(m)} + S^{(m)}$, where $L^{(m)}$ is low-rank and $S^{(m)}$ is sparse diagonal noise; the sample covariance $\widehat\Sigma^{(m)}$ is the observation matrix. Source: $p_1=10, r_1=3, n_1=500$; target: $p_2=50, r_2=4$ ($\delta_{r,2}=1$), with $n_2\in\{30,50,80,100,120,150,200,250,300\}$. The target low-rank component embeds $L^{(1)}$ into $\mathbb R^{p_2\times p_2}$ plus a rank-one innovation, and the sparse component analogously combines embedded source sparsity with a few diagonal edits. We compare \textbf{Transfer AltProj} (Algorithm~\ref{alg:transfer-altproj} with anchors from $\widehat\Sigma^{(1)}$), a non-transfer structured estimator, and single-task PCA, measured in Frobenius error of $\widehat L^{(2)}$ and $\sin(\Theta)$-distance between $\widehat U^{(2)}$ and $U^{(2)}$ over $50$ trials. Theoretical analysis appears in Appendix~\ref{appendix:pca_application} and additional details in Appendix~\ref{append:experiment_details}.

\begin{figure}[ht]
\centering
\includegraphics[width=0.88\columnwidth]{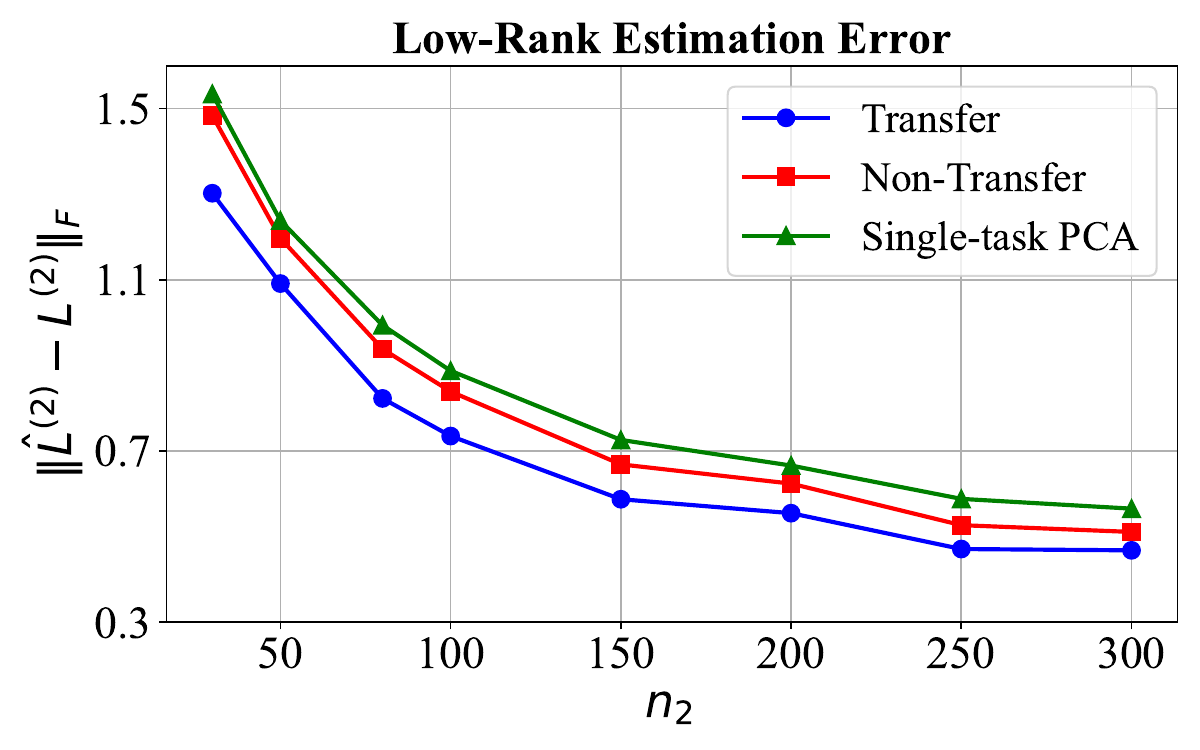}
\includegraphics[width=0.88\columnwidth]{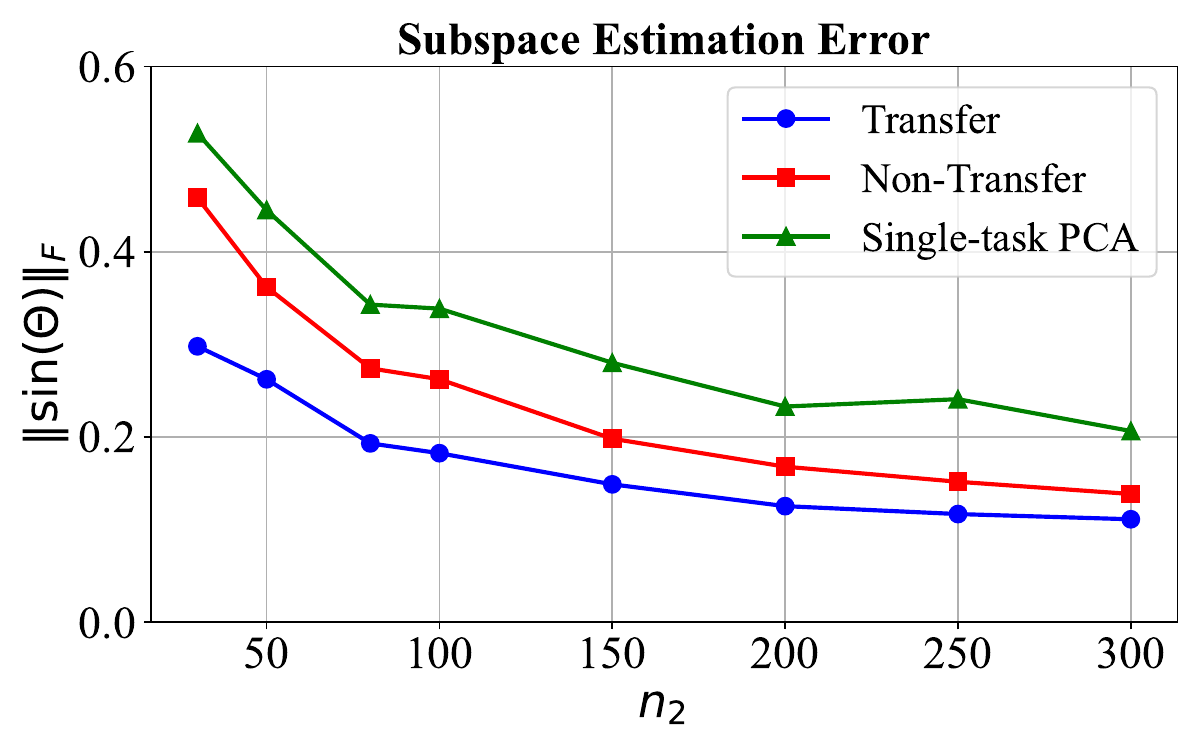}
\caption{Synthetic covariance estimation under enlarged dimension. Curves are means over 50 trials with $\pm 1$ standard-error shading. Transfer AltProj consistently outperforms Non-Transfer and single-task PCA across all $n_2$, with the largest gains in the low-sample regime, consistent with error scaling in $\delta_{r,2}=1$ rather than $r_2=4$.}
\label{fig:L_error_n2}
\end{figure}

\subsection{NYC Taxi Markov Transitions: Transfer under Real Ambient Growth}
\label{sec:empirical_real}

The synthetic study isolates the mechanism, but the central empirical question is whether the framework provides gains on real data not engineered to satisfy the transfer model. We test this on NYC taxi trip records (January 2024, $\sim$3M trips), treating each trip as a one-step Markov transition between districts. The target is the full $p_2 = 255$-district transition matrix; the source is constructed by randomly selecting a $p_1 = 200$-district subnetwork, so the remaining $\delta_{p,2} = 55$ districts represent genuinely new state-space directions on which the source provides no information. For each of $20$ trials we draw a single source and a single target trajectory, learn source anchors via Algorithm~\ref{alg:transfer-altproj} on $\widehat F^{(1)}$, and estimate $P^{(2)\star}$ using \textbf{Transfer AltProj}, a \textbf{non-transfer structured estimator} (Algorithm~\ref{alg:transfer-altproj} without anchoring), and \textbf{truncated SVD}. Performance is the principal-angle $\sin(\Theta)$ distance between estimated and ground-truth left singular subspaces of $F^{(2)\star}$; preprocessing and protocol details are in Appendix~\ref{appendix:real_data}.

Transfer AltProj attains mean error $\mathbf{0.915}$ ($\pm 0.07$), against $1.679$ ($\pm 0.09$) for the non-transfer estimator and $1.703$ ($\pm 0.09$) for truncated SVD ($\pm 1$ standard deviation across 20 trials), which is approximately a $45\%$ reduction relative to both baselines. This dataset was not engineered to fit the transfer model: the partition into inherited and new districts is random, so any rotation, partial overlap, or non-stationarity in the underlying transition process can only hurt the transfer estimator. The substantial gain on this real, noisy data therefore directly supports the framework's practical relevance beyond the synthetic setting of Section~\ref{sec:empirical_synthetic}.

\section{Conclusion}

We proposed a transfer framework for structured matrix estimation under growing ambient and representation dimensions, with an anchored estimator and deterministic error guarantees that scale with incremental rather than full target complexity. Applications to Markov transition estimation and empirical results on synthetic and real NYC taxi data demonstrate consistent transfer benefits.



\bibliographystyle{plainnat}
\bibliography{main}

\clearpage
\appendix

\section{Algorithms}\label{append:algo}

\begin{algorithm}[h]
\caption{\textsc{AnchoredSpectralUpdate}: $L_{t+1} \leftarrow \mathsf P_{\text{anch-rank}}(M_{t+1};\widetilde U^{(1)},\widetilde V^{(1)},\delta_{r,2},r_1)$}
\label{alg:anchored-lowrank-proj}
\begin{algorithmic}[1]
\REQUIRE Matrix $M_{t+1}\in\mathbb{R}^{p_2\times q_2}$; anchored subspaces $\widetilde U^{(1)}\in\mathbb{R}^{p_2\times r_1}$, $\widetilde V^{(1)}\in\mathbb{R}^{q_2\times r_1}$; rank increment $\delta_{r,2}$.
\ENSURE Anchored low-rank estimate $L_{t+1}\in\mathbb{R}^{p_2\times q_2}$.

\STATE $P_{\widetilde U} \leftarrow \widetilde U^{(1)}\widetilde U^{(1)\top}$,\quad $P_{\widetilde V} \leftarrow \widetilde V^{(1)}\widetilde V^{(1)\top}$.
\STATE $M^{\perp}_{t+1} \leftarrow (I-P_{\widetilde U})\,M_{t+1}\,(I-P_{\widetilde V})$.
\STATE Compute rank-$\delta_{r,2}$ SVD within the two orthogonal complements (use empty innovation frames if $\delta_{r,2}=0$):
$M^{\perp}_{t+1} \approx U_{\Delta,t+1}\Sigma_{\Delta,t+1}V_{\Delta,t+1}^\top$.
\STATE Complete innovation frames orthonormally within the complements if the residual rank is smaller than $\delta_{r,2}$.
\STATE $U^{(2)}_{t+1} \leftarrow [\,\widetilde U^{(1)}\;\; U_{\Delta,t+1}\,]$,\quad
       $V^{(2)}_{t+1} \leftarrow [\,\widetilde V^{(1)}\;\; V_{\Delta,t+1}\,]$.
\STATE $A^{(2)}_{t+1} \leftarrow (U^{(2)}_{t+1})^\top M_{t+1} V^{(2)}_{t+1}$.
\STATE $L_{t+1} \leftarrow U^{(2)}_{t+1} A^{(2)}_{t+1} (V^{(2)}_{t+1})^\top$.
\STATE \textbf{return} $L_{t+1}$.
\end{algorithmic}
\end{algorithm}

\begin{algorithm}[h]
\caption{\textsc{SparseEditProj}: $S \leftarrow \mathsf P_{\mathrm{edit}}(M;S_0,\delta_{s,2})$}
\label{alg:sparse-edit-proj}
\begin{algorithmic}[1]
\REQUIRE Matrix $M\in\mathbb{R}^{p_2\times q_2}$; sparse anchor $S_0\in\mathbb{R}^{p_2\times q_2}$; edit budget $\delta_{s,2}$.
\ENSURE Sparse-edit projection $S\in\mathbb{R}^{p_2\times q_2}$.

\STATE $R \leftarrow M - S_0$.
\STATE $E \leftarrow \mathcal{H}_{\delta_{s,2}}(R)$ \COMMENT{Keep the $\delta_{s,2}$ largest-magnitude entries of $R$, zero out the rest.}
\STATE $S \leftarrow S_0 + E$.
\STATE \textbf{return} $S$.
\end{algorithmic}
\end{algorithm}

\section{Proof of Theorem~\ref{thm:main}}\label{appendix:proof_main}

\begin{lemma}[Regularity of the source-adjusted comparison point]
\label{lem:comparison-regularity}
Write $d=\delta_{r,2}$ and $r=r_1+d=r_2$. Let
\[
U_0=B(U^{(1)}),\qquad V_0=B(V^{(1)}),\qquad
U=[U_0\ U_\Delta],\qquad V=[V_0\ V_\Delta],
\]
where $U,V$ have orthonormal columns, and suppose
$L^{(2)}=UA^{(2)}V^\top$ has rank $r$.
Let $\widetilde U^{(1)},\widetilde V^{(1)}$ be the embedded
source estimates, each with orthonormal columns. Choose
\[
\bar P_U\in\arg\min_{P\in\mathcal O_{r_1,r_1}}
\|U_0P-\widetilde U^{(1)}\|_F,\qquad
\bar P_V\in\arg\min_{P\in\mathcal O_{r_1,r_1}}
\|V_0P-\widetilde V^{(1)}\|_F.
\]
Assume that, for some $\mu_0>0$,
\begin{equation}
\label{eq:comparator-row-regularity}
\begin{aligned}
\max\{\|U\|_{2,\infty},\|\widetilde U^{(1)}\|_{2,\infty}\}
&\le \mu_0\sqrt{r/p_2},\\
\max\{\|V\|_{2,\infty},\|\widetilde V^{(1)}\|_{2,\infty}\}
&\le \mu_0\sqrt{r/q_2},
\end{aligned}
\end{equation}
and
\begin{equation}
\label{eq:comparator-source-proximity}
\max\bigl\{
\|U_0\bar P_U-\widetilde U^{(1)}\|_2,
\|V_0\bar P_V-\widetilde V^{(1)}\|_2
\bigr\}\le \tfrac12.
\end{equation}
Define
\[
P_U=\operatorname{Diag}(\bar P_U,I_d),\qquad
P_V=\operatorname{Diag}(\bar P_V,I_d),
\]
\[
T_U=[\widetilde U^{(1)}\ U_\Delta],\qquad
T_V=[\widetilde V^{(1)}\ V_\Delta],\qquad
\widetilde A=P_U^\top A^{(2)}P_V,\qquad
\widetilde L=T_U\widetilde A T_V^\top.
\]
Then $\widetilde L$ has rank $r$, its column and row spaces contain
$\operatorname{span}(\widetilde U^{(1)})$ and
$\operatorname{span}(\widetilde V^{(1)})$, respectively, and
\[
\widetilde L\in\mathcal M_{2\mu_0}.
\]
Moreover, writing
\[
\alpha_U=\|U_0\bar P_U-\widetilde U^{(1)}\|_F,\qquad
\alpha_V=\|V_0\bar P_V-\widetilde V^{(1)}\|_F,
\]
we have
\begin{equation}
\label{eq:comparator-approximation-error}
\|L^{(2)}-\widetilde L\|_F^2
\le 3(\alpha_U^2+\alpha_V^2)\|A^{(2)}\|_2^2.
\end{equation}
Consequently, if $\widetilde S=S_0+\Delta_S^{(2)}$ and
$\|\Delta_S^{(2)}\|_0\le\delta_{s,2}$, then
$(\widetilde L,\widetilde S)$ is feasible for
\eqref{eqn:optimization-problem} with sparse-edit constraint
$\|S-S_0\|_0\le\delta_{s,2}$ and fitting incoherence budget
$\mu\ge 2\mu_0$.
\end{lemma}

\begin{proof}
Set $C_U=\widetilde U^{(1)\top}U_\Delta$. Since
$U_0^\top U_\Delta=0$, \eqref{eq:comparator-source-proximity} gives
\[
\|C_U\|_2
=\|(\widetilde U^{(1)}-U_0\bar P_U)^\top U_\Delta\|_2
\le\tfrac12.
\]
Thus
\[
T_U^\top T_U=
\begin{bmatrix}I_{r_1}&C_U\\ C_U^\top&I_d\end{bmatrix},
\qquad
\tfrac12 I_r\preceq T_U^\top T_U\preceq\tfrac32 I_r.
\]
In particular, $T_U$ has full column rank and
$Q_U=T_U(T_U^\top T_U)^{-1/2}$ is an orthonormal basis of its
column space. Since $U_\Delta$ is a column block of $U$,
\eqref{eq:comparator-row-regularity} implies
\[
\|T_U\|_{2,\infty}^2
\le \|\widetilde U^{(1)}\|_{2,\infty}^2
   +\|U_\Delta\|_{2,\infty}^2
\le 2\mu_0^2r/p_2.
\]
Therefore,
\[
\|Q_U\|_{2,\infty}
\le \|T_U\|_{2,\infty}
   \|(T_U^\top T_U)^{-1/2}\|_2
\le 2\mu_0\sqrt{r/p_2}.
\]
The same argument gives a full-column-rank $T_V$ and an orthonormal
basis $Q_V$ satisfying
$\|Q_V\|_{2,\infty}\le 2\mu_0\sqrt{r/q_2}$.

Because $L^{(2)}$ has rank $r$, both $A^{(2)}$ and
$\widetilde A$ are nonsingular. Hence $\widetilde L$ has rank $r$,
with column and row spaces equal to those of $T_U$ and $T_V$.
Its singular-vector bases differ from $Q_U,Q_V$ only by orthogonal
transformations, which preserve row norms. This proves
$\widetilde L\in\mathcal M_{2\mu_0}$.
These spaces contain the estimated source anchors, so they also admit
orthonormal bases beginning with $\widetilde U^{(1)}$ and
$\widetilde V^{(1)}$, with $d$ orthogonal innovation columns.
The unrestricted coefficient matrix absorbs this change of basis,
without changing $\widetilde L$. This establishes the required
anchored parameterization.

For the approximation error, observe that
\[
UP_U-T_U=[U_0\bar P_U-\widetilde U^{(1)}\ 0],\qquad
VP_V-T_V=[V_0\bar P_V-\widetilde V^{(1)}\ 0].
\]
Since $L^{(2)}=UP_U\widetilde A(VP_V)^\top$,
\[
\begin{aligned}
L^{(2)}-\widetilde L
={}&(UP_U-T_U)\widetilde A(VP_V)^\top\\
&+T_U\widetilde A(VP_V-T_V)^\top.
\end{aligned}
\]
Using $\|VP_V\|_2=1$, $\|T_U\|_2\le\sqrt{3/2}\le\sqrt2$,
and $\|\widetilde A\|_2=\|A^{(2)}\|_2$, we obtain
\[
\|L^{(2)}-\widetilde L\|_F
\le (\alpha_U+\sqrt2\alpha_V)\|A^{(2)}\|_2.
\]
Squaring and applying Cauchy--Schwarz proves
\eqref{eq:comparator-approximation-error}.
Finally, $\|\widetilde S-S_0\|_0
=\|\Delta_S^{(2)}\|_0\le\delta_{s,2}$, proving feasibility of
both components.
\end{proof}

\subsection{Separation and transfer-adapted noise control}

\begin{lemma}[A rectangular separation margin]\label{lem:rectangular-separation}
Let $L,L'\in\mathbb R^{p\times q}$ admit rank-budget-$r$ singular-vector
frames satisfying the definition of $\mathcal M_\mu$ in the main text.
If $D=L-L'$ and $H$ has at most $2s$ nonzero entries, then
\[
|\langle D,H\rangle|
\le \sqrt{\frac{2c_{\mathrm{sep}}\mu^2r^3s}{\min\{p,q\}}}
\,\|D\|_F\|H\|_F,
\]
where $c_{\mathrm{sep}}$ is universal. In particular, if
$s\le\min\{p,q\}/(8c_{\mathrm{sep}}\mu^2r^3)$, then
\[
\|D+H\|_F^2\ge\tfrac12(\|D\|_F^2+\|H\|_F^2).
\]
\end{lemma}
\begin{proof}
Pad both matrices with zeros to dimension $N=\max\{p,q\}$.
Their padded singular-vector frames satisfy the square-matrix
incoherence convention of \citet{chai2025structured} with
parameter $\mu_{\mathrm{CF}}=\mu^2N/\min\{p,q\}$.
That lemma, which permits zero singular values within the rank budget,
gives
\[
\|D\|_{\max}^2\le
\frac{c_{\mathrm{sep}}\mu^2r^3}{\min\{p,q\}}\|D\|_F^2.
\]
Padding preserves these norms. Since
$\|H\|_{1,1}\le\sqrt{2s}\|H\|_F$, entrywise duality proves
the first inequality. Under the stated restriction, its coefficient
is at most $1/2$, so
$|\langle D,H\rangle|\le(\|D\|_F^2+\|H\|_F^2)/4$.
Expand $\|D+H\|_F^2$ to finish the proof.
\end{proof}

\begin{lemma}[Noise control with estimated anchors]\label{lem:anchored-noise}
Let $U_a\in\mathbb R^{p\times r_1}$ and
$V_a\in\mathbb R^{q\times r_1}$ have orthonormal columns and let
$P=U_aU_a^\top$, $Q=V_aV_a^\top$.
Suppose both $L_1,L_2$ satisfy
\[
\operatorname{rank}((I-P)L_j)\le d,\qquad
\operatorname{rank}(L_j(I-Q))\le d\quad (j=1,2).
\]
For $D=L_1-L_2$ and any $W$,
\[
|\langle W,D\rangle|
\le \left\{r_1\|U_a^\top WV_a\|_2^2+6d\|W\|_2^2\right\}^{1/2}
\|D\|_F.
\]
\end{lemma}
\begin{proof}
Decompose the matrix error, rather than its original coefficient blocks:
\[
D_{11}=PDQ,\quad D_{12}=PD(I-Q),\quad
D_{21}=(I-P)DQ,\quad D_{22}=(I-P)D(I-Q).
\]
These four blocks are mutually Frobenius-orthogonal, because either their
left or their right factors lie in orthogonal projector ranges. Thus
\begin{equation}\label{eq:orthogonal_error}
\|D\|_F^2=\sum_{i,j=1}^2\|D_{ij}\|_F^2.
\end{equation}
The inherited block has rank at most $r_1$; each other block has rank
at most $2d$ by the two rank assumptions and rank subadditivity.
Let $C=U_a^\top D V_a$. Then
$D_{11}=U_a C V_a^\top$ and $\|C\|_F=\|D_{11}\|_F$.
Nuclear--spectral duality and
$\|M\|_*\le\sqrt{\operatorname{rank}(M)}\|M\|_F$ give
\[
\begin{aligned}
|\langle W,D\rangle|
\le{}&\sqrt{r_1}\|U_a^\top WV_a\|_2\|D_{11}\|_F\\
&+\sqrt{2d}\|W\|_2
(\|D_{12}\|_F+\|D_{21}\|_F+\|D_{22}\|_F).
\end{aligned}
\]
Cauchy--Schwarz across the four blocks and
\eqref{eq:orthogonal_error} prove the claim. No orthogonality between
the population innovations and the estimated anchors is needed.
\end{proof}

\subsection{Deterministic bound and source specialization}
\begin{proof}[Proof of Theorem~\ref{thm:main}]
Suppress the task superscript and use the comparison matrix
$L^\circ=\widetilde L$ from Lemma~\ref{lem:comparison-regularity}.
Set
\[
S^\circ=S_0+\Delta_S^{(2)},\quad
D=\widehat L-L^\circ,\quad H=\widehat S-S^\circ,
\]
\[
E_L=L-L^\circ,\quad E_S=S-S^\circ,\quad E=E_L+E_S.
\]
The comparison pair is feasible at the fitting budget $\mu=2\mu_0$.
Moreover, $H$ is the difference of two edits of size at most
$\delta_{s,2}$, and hence $\|H\|_0\le2\delta_{s,2}$.
Feasible $\varepsilon_{\mathrm{opt}}$-optimality gives
\[
\tfrac12\|Y-\widehat L-\widehat S\|_F^2
\le\tfrac12\|Y-L^\circ-S^\circ\|_F^2+
\varepsilon_{\mathrm{opt}}.
\]
Since $Y=L^\circ+S^\circ+W+E$, expansion yields
\begin{equation}\label{eq:basic-comparison-inequality}
\tfrac12\|D+H\|_F^2
\le\langle W,D+H\rangle+\langle E,D+H\rangle+
\varepsilon_{\mathrm{opt}}.
\end{equation}
Both fitted and comparison low-rank matrices belong to $\mathcal M_\mu$,
so Lemma~\ref{lem:rectangular-separation} applies.
Let $x^2=\|D\|_F^2+\|H\|_F^2$ and
$b^2=\|E_L\|_F^2+\|E_S\|_F^2$.
The left-hand side of \eqref{eq:basic-comparison-inequality} is at least
$x^2/4$, and $|\langle E,D+H\rangle|\le2bx$.

For $P=P_{\widetilde U}$ and $Q=P_{\widetilde V}$, both low-rank
matrices have at most $\delta_{r,2}$ directions outside the estimated
anchors on each side. For the comparison, this follows directly from
$(I-P)T_U=[0\ (I-P)U_\Delta]$ and its right-hand analogue;
nonorthogonality of these columns is immaterial to their rank.
Lemma~\ref{lem:anchored-noise} and entrywise duality therefore imply
\[
|\langle W,D+H\rangle|\le Kx,
\]
where
\[
K^2=r_1\|\widetilde U^{(1)\top}W\widetilde V^{(1)}\|_2^2
+6\delta_{r,2}\|W\|_2^2+2\delta_{s,2}\|W\|_{\max}^2.
\]
It follows that $x^2/4\le(K+2b)x+\varepsilon_{\mathrm{opt}}$.
Young's inequality gives
\[
x^2\le8(K+2b)^2+8\varepsilon_{\mathrm{opt}}
\lesssim K^2+b^2+\varepsilon_{\mathrm{opt}}.
\]
Finally,
\[
\|\widehat L-L\|_F^2+\|\widehat S-S\|_F^2\le2x^2+2b^2,
\]
and Lemma~\ref{lem:comparison-regularity} bounds
$\|E_L\|_F^2\le3(\alpha_U^2+\alpha_V^2)\|A\|_2^2$.
Because zero-padding is an isometry,
$\|E_S\|_F^2=\|\widehat S^{(1)}-S^{(1)}\|_F^2$.
Combining these facts proves \eqref{eq:main-transfer-bound}.
\end{proof}

\begin{proof}[Proof of Corollary~\ref{thm:chai2025_(1)}]
Let $\Delta_1=\widehat L^{(1)}-L^{(1)}$.
Wedin's singular-subspace perturbation bound
\citep{chen2021spectral}, together with the relationship between
Procrustes and sine distances, gives
\[
\alpha_U^2+\alpha_V^2
\lesssim \frac{\|\Delta_1\|_F^2}
{(\sigma_{r_1}-\|\Delta_1\|_2)^2}
\lesssim R_1/\sigma_{r_1}^2.
\]
The last step uses the stated signal-separation condition.
Substitute this and the source sparse-error bound into
Theorem~\ref{thm:main}. All source-anchor row-regularity conditions are
retained; the source norm bound alone does not imply them.
\end{proof}

\paragraph{Feasibility and existence.}
Under the theorem's assumptions the comparison lemma makes the feasible
set nonempty. It is closed: anchored orthonormal frames and the
incoherent singular-vector frames range over compact sets, and the
cardinality constraint is closed. For any bounded-objective sequence,
separation relative to the feasible comparison pair bounds both
$\|L-L^\circ\|_F$ and $\|S-S^\circ\|_F$.
A convergent subsequence therefore attains the infimum.
This is an existence argument for the statistical estimator, not an
algorithmic convergence guarantee.

\paragraph{Compatibility of embedding and incoherence.}
An $r_1$-dimensional orthonormal source frame supported on $p_1$ inherited
rows has squared maximum row norm at least $r_1/p_1$.
Thus the row-regularity condition requires
$\mu_0^2\ge r_1p_2/(r_2p_1)$, and analogously on the right.
Large expansion may therefore require a larger incoherence parameter
and a smaller sufficient edit budget. The numerical studies evaluate
the practical procedure and are not claimed to satisfy every
conservative sufficient condition in this theorem.

\section{Robustness to Approximate Anchoring}\label{appendix:robust}

Here we replace, rather than supplement, the exact inheritance assumption.
Let the true rank-$r_2$ target factors be
\[
U^{(2)}=[U_a^{(2)}\ U_\Delta],\qquad
V^{(2)}=[V_a^{(2)}\ V_\Delta],
\]
with orthonormal columns, $r_1$ inherited columns, and
$\delta_{r,2}$ innovation columns. Suppose
\begin{equation}\label{eq:approx-anchor}
\begin{aligned}
\min_{R\in\mathcal O_{r_1,r_1}}
\|B(U^{(1)})R-U_a^{(2)}\|_F&\le\epsilon_U,\\
\min_{R\in\mathcal O_{r_1,r_1}}
\|B(V^{(1)})R-V_a^{(2)}\|_F&\le\epsilon_V.
\end{aligned}
\end{equation}
Retain the row-regularity condition for the \emph{true} target factors
and the estimated source anchors, the sparse-edit model, and the
fitting budget $\mu=2\mu_0$.
The quantities $\alpha_U,\alpha_V,e_U,e_V$ continue to measure error
between the population source frames and their estimates, as in the
main text. Require the combined proximity condition
\begin{equation}\label{eq:combined-anchor-proximity}
\max\{e_U+\epsilon_U,e_V+\epsilon_V\}\le\tfrac12.
\end{equation}

\begin{theorem}[Robust deterministic transfer bound]\label{thm:main-robust}
Under the preceding approximate-inheritance conditions, the same
statistical feasible set and approximate-optimality condition give
\[
\begin{aligned}
&\|\widehat L^{(2)}-L^{(2)}\|_F^2+
\|\widehat S^{(2)}-S^{(2)}\|_F^2\\
&\quad\lesssim \mathcal E_{\mathrm{tgt}}+\mathcal E_{\mathrm{src}}
 +(\epsilon_U^2+\epsilon_V^2)\|A^{(2)}\|_2^2
 +\varepsilon_{\mathrm{opt}}.
\end{aligned}
\]
The definitions of $\mathcal E_{\mathrm{tgt}}$ and
$\mathcal E_{\mathrm{src}}$ are those in Section~\ref{sec:theory}.
\end{theorem}

\begin{proof}
Let $R_U^\epsilon$ attain the first minimum in
\eqref{eq:approx-anchor}, and put
$R_U^\circ=(R_U^\epsilon)^\top\bar P_U$.
Then
\[
\begin{aligned}
\|U_a^{(2)}R_U^\circ-\widetilde U^{(1)}\|_F
&\le\epsilon_U+\alpha_U,\\
\|U_a^{(2)}R_U^\circ-\widetilde U^{(1)}\|_2
&\le\epsilon_U+e_U.
\end{aligned}
\]
The analogous construction gives $R_V^\circ$ and the corresponding
right-hand bounds. Apply the construction in
Lemma~\ref{lem:comparison-regularity} with $U_a^{(2)},V_a^{(2)}$ in
place of the exact population source anchors and with these rotations.
The lemma's proof needs only the displayed distances, not optimality
of the rotations themselves. Condition~\eqref{eq:combined-anchor-proximity}
ensures that the comparison Gram matrices remain bounded away from
singularity. The same row-norm calculation establishes feasibility in
$\mathcal M_{2\mu_0}$, and
\[
\begin{aligned}
\|L^{(2)}-L^\circ\|_F^2
&\le3\{(\alpha_U+\epsilon_U)^2+
       (\alpha_V+\epsilon_V)^2\}\|A^{(2)}\|_2^2\\
&\le6(\alpha_U^2+\alpha_V^2+\epsilon_U^2+\epsilon_V^2)
\|A^{(2)}\|_2^2.
\end{aligned}
\]
Keep $S^\circ=S_0+\Delta_S^{(2)}$.
The proof of Theorem~\ref{thm:main}, starting from
\eqref{eq:basic-comparison-inequality}, applies verbatim to this
feasible comparison pair and proves the result.
\end{proof}

\paragraph{Why the residual is quadratic in anchoring error.}
For aligned full comparison frames $T_U,T_V$ and the correspondingly
rotated full coefficient matrix $A^\circ$, the residual is exactly
\[
L^{(2)}-T_UA^\circ T_V^\top
=(U^{(2)}R_U-T_U)A^\circ(V^{(2)}R_V)^\top
+T_UA^\circ(V^{(2)}R_V-T_V)^\top,
\]
where $R_U,R_V$ are the block-diagonal rotations used in the construction.
The bound follows from the Frobenius factor discrepancies and uniformly
bounded operator norms of the frames. There is no multiplication of an
$r_2$-column frame by an $r_1\times r_1$ coefficient block, nor any need
to add the innovation twice. In the basic inequality this residual is
controlled directly by Cauchy--Schwarz and Young's inequality;
substituting it indiscriminately into spectral-noise terms could instead
introduce unnecessary rank factors.

\paragraph{Interpretation.}
Population mismatch and source-estimation error have the same
approximation geometry but distinct origins. More source observations
can reduce the latter, not the former. The additional term need not be
dominated by source error: it is lower order only under an explicit
comparison with the relevant statistical-error scale.
The theorem controls approximate inheritance, not arbitrary drift,
and does not certify the numerical solver.

For equal-dimensional orthonormal anchor frames $U_0,U_a$, the
Procrustes and principal-angle distances satisfy
\[
\|\sin\Theta(U_0,U_a)\|_F^2
\le\min_R\|U_0R-U_a\|_F^2
\le2\|\sin\Theta(U_0,U_a)\|_F^2.
\]
These inequalities follow from $\sin^2\theta\le2(1-\cos\theta)
\le2\sin^2\theta$ on $[0,\pi/2]$.
Small angle alone does not bound the number of nonzero orthogonal
innovation directions. Conversely, in the approximate model many weak
deviations can be left as approximation error rather than represented
exactly by additional directions; this is a bias--complexity tradeoff,
not an implication that geometric closeness forces small exact rank.

\section{Application to Markov Transition Matrix Estimation}
\label{appendix:markov_application}

\begin{assumption}[Markov observations and structural transfer]
\label{asp:markov-structured}
The target chain on $[p_2]$ is irreducible and aperiodic, with stationary
distribution $\pi^{(2)}$ and initial state distributed as $\pi^{(2)}$.
Its frequency matrix has the exact structured transfer decomposition
and satisfies Assumption~\ref{asp:incoherence-sparsity} together with
the source estimates on which we condition. The source observations
are independent of the target trajectory. Target estimation means a
feasible $\varepsilon_{\mathrm{opt}}$-optimal solution of
\eqref{eqn:optimization-problem}; it does not mean an uncertified
output of Algorithm~\ref{alg:transfer-altproj}.
\end{assumption}

\subsection{Concentration including the inherited block}
We provide the concentration argument explicitly, rather than applying
an independent-entry noise bound to a Markov frequency matrix.
Suppress the task superscript, and let $n$ be the number of observed
transitions. Write $\pi_{\max}=\max_i\pi_i$,
$p_{\max}=\max_{i,j}P_{ij}$, and
$d_X(t)=\max_x\|P^t(x,\cdot)-\pi\|_{\mathrm{TV}}$.
For integer mixing time $\tau_\star$ satisfying $d_X(\tau_\star)\le1/4$,
submultiplicativity of the pairwise total-variation distance implies
\begin{equation}\label{eq:mixing-decay}
d_X(k\tau_\star)\le2^{-k}\quad(k\ge1)
\end{equation}
\citep{levin2017markov}. Indeed, if
$\bar d(t)=\max_{x,y}\|P^t(x,\cdot)-P^t(y,\cdot)\|_{\mathrm{TV}}$,
then $d_X(t)\le\bar d(t)\le2d_X(t)$ and
$\bar d(s+t)\le\bar d(s)\bar d(t)$.

\begin{lemma}[Full, entrywise, and compressed frequency noise]
\label{lem:markov-noise}
Let $U_a,V_a\in\mathbb R^{p\times r_1}$ be fixed orthonormal frames
independent of the target trajectory, and set
$a=\max\{\|U_a\|_{2,\infty}^2,\|V_a\|_{2,\infty}^2\}$.
For $b$ and $h_n$ as in Section~\ref{sec:transition-matrix-estimation}
and $n\ge2b$, with probability at least $1-\eta$ the empirical frequency
noise $W=\widehat F-F$ simultaneously satisfies
\[
\begin{aligned}
\|W\|_2^2&\lesssim\pi_{\max}h_n+h_n^2,\\
\|W\|_{\max}^2&\lesssim\pi_{\max}p_{\max}h_n+h_n^2,\\
\|U_a^\top WV_a\|_2^2&\lesssim\pi_{\max}a h_n+a^2h_n^2.
\end{aligned}
\]
The same bounds hold conditional on independently estimated source frames.
\end{lemma}

\begin{proof}
\textbf{Blocking and coupling.}
Partition transition times $0,\ldots,n-1$ into $b$ residue classes
modulo $b$. Within one class, consecutive pairs
$(X_t,X_{t+1})$ and $(X_{t+b},X_{t+b+1})$ are separated by $b-1$
chain steps. Conditional on all preceding pairs in that class, the
next pair's distribution differs from the stationary pair law
$F_{ij}=\pi_iP_{ij}$ in total variation by at most $d_X(b-1)$.
The Markov property and contraction of total variation under the
transition from a state to a state pair justify this bound.
Successive maximal couplings, or a telescoping comparison of the
conditional laws, show that a class containing $m$ pairs differs
from $m$ independent stationary pairs by total variation at most
$(m-1)d_X(b-1)$.
By \eqref{eq:mixing-decay} and the choice of $b$,
$d_X(b-1)\le\eta/(8n)$.
A union bound over the residue classes therefore costs at most $\eta/8$.
The classes need not be independent of each other.

\textbf{Independent-pair variance bounds.}
For one stationary pair $(I,J)$, put $Y=e_Ie_J^\top$.
Then $\mathbb EY=F$, $\|Y-F\|_2\le2$, and
\[
\mathbb E[(Y-F)(Y-F)^\top]
=\operatorname{Diag}(\pi)-FF^\top
\preceq\pi_{\max}I,
\]
with the same bound for the transposed product, by stationarity.
Each entry indicator has variance at most
$F_{ij}\le\pi_{\max}p_{\max}$ and centered magnitude at most one.
For $Z=U_a^\top YV_a=u_Iv_J^\top$, where
$u_i=U_a^\top e_i$ and $v_j=V_a^\top e_j$, we have
$\|Z-\mathbb EZ\|_2\le2a$ and
\[
\begin{aligned}
\mathbb E[ZZ^\top]
&=\sum_{i,j}\pi_iP_{ij}\|v_j\|_2^2u_i u_i^\top
\preceq a\,U_a^\top\operatorname{Diag}(\pi)U_a
\preceq a\pi_{\max}I,\\
\mathbb E[Z^\top Z]&\preceq a\pi_{\max}I.
\end{aligned}
\]
Centering subtracts a positive-semidefinite matrix from each variance.

\textbf{Bernstein bounds and recombination.}
For a residue class with $m$ independent pairs, rectangular matrix
Bernstein \citep{tropp2012userfriendly} bounds a centered sum by
$C\{\sqrt{m\nu\ell}+R\ell\}$, except on an event of probability
at most a matrix-dimension factor times $e^{-\ell}$.
Here $(\nu,R)$ can be $(\pi_{\max},2)$ for $Y-F$,
$(a\pi_{\max},2a)$ for $Z-\mathbb EZ$, or
$(\pi_{\max}p_{\max},1)$ for an entry.
Take $\ell=\log(32p^2b/\eta)$ and union bound over all classes,
the two matrix events, and all $p^2$ entry events. Their total failure
probability, together with coupling, is less than $\eta$.
If $m_j$ are the class sizes, then $\sum_jm_j=n$ and
$\sum_j\sqrt{m_j}\le\sqrt{bn}$.
Summing the class bounds and dividing by $n$ yields
\[
\begin{aligned}
\|W\|_2&\lesssim\sqrt{\pi_{\max}h_n}+h_n,\\
\|W\|_{\max}&\lesssim\sqrt{\pi_{\max}p_{\max}h_n}+h_n,\\
\|U_a^\top WV_a\|_2&\lesssim\sqrt{a\pi_{\max}h_n}+a h_n.
\end{aligned}
\]
Squaring proves the result. Independence of source and target samples
allows the entire argument to be conditioned on the source frames.
\end{proof}

\begin{proof}[Proof of Corollary~\ref{cor:transfer-markov}]
Apply Lemma~\ref{lem:markov-noise} with
$U_a=\widetilde U^{(1)}$, $V_a=\widetilde V^{(1)}$ and substitute
its three bounds into Theorem~\ref{thm:main}.
The total frequency error is at most twice the sum of squared
component errors. This gives the stated rate.
If source regularity holds with probability at least $1-\eta_1$,
the conditional statement gives an unconditional joint success
probability at least $1-\eta-\eta_1$; we do not assume this source
event occurs automatically.
The approximate-anchoring version follows from
Theorem~\ref{thm:main-robust} with the additional mismatch term retained.
\end{proof}

\paragraph{Target-only benchmark.}
For the target-only constrained estimator with the same rank budget and
full sparsity budget $s_2$, require the corresponding full-sparsity
separation condition. The basic inequality, rank at most $2r_2$ for the
low-rank error, and support at most $2s_2$ for the sparse error give
\[
\|\widehat L_{\mathrm{base}}-L\|_F^2+
\|\widehat S_{\mathrm{base}}-S\|_F^2
\lesssim r_2\|W\|_2^2+s_2\|W\|_{\max}^2+
\varepsilon_{\mathrm{base}}.
\]
Lemma~\ref{lem:markov-noise} gives
$\pi_{\max}h_n(r_2+s_2p_{\max})+h_n^2(r_2+s_2)$ plus
optimization error. This uses the frequency-noise scale of
\citet{chai2025structured,zhang2019spectral}, not an additional
ambient factor $p_2$ multiplying their rank term.
Our blocking argument retains explicit Bernstein remainders and is
not asserted to be mixing- or logarithm-optimal.
If $h_n\le\pi_{\max}p_{\max}$, all quadratic remainders are
absorbed in their linear terms. In particular, since $a\le1$,
$a^2h_n^2\le a\pi_{\max}h_n$.
The inherited coefficient is $r_1a$, which may be of order $r_1^2/p_2$
for sufficiently spread-out anchors. Neither such spreading nor a
negligible source error is automatic under arbitrary expansion.

\subsection{Row normalization}
\begin{lemma}[From frequency error to transition error]
\label{lem:frequency-normalization}
Let $F=\operatorname{Diag}(\pi)P$ with $P$ row-stochastic and
$\pi_{\min}>0$. For any estimate $\widehat F$ write
$D_F=\widehat F-F$ and $\widehat\pi=\widehat F\mathbf1$.
Suppose $\min_i\widehat\pi_i\ge\pi_{\min}/2$ and define
$Q=\operatorname{Diag}(\widehat\pi)^{-1}\widehat F$.
For its rowwise Euclidean simplex projection $\widehat P$,
\[
\begin{aligned}
\|\widehat P-P\|_F^2
&\le\frac{8}{\pi_{\min}^2}
\left\{\|D_F\|_F^2+
\sum_i(D_F\mathbf1)_i^2\|P_{i,:}\|_2^2\right\}\\
&\le\frac{8}{\pi_{\min}^2}
\{\|D_F\|_F^2+p_{\max}\|D_F\mathbf1\|_2^2\}.
\end{aligned}
\]
Consequently,
$\|\widehat P-P\|_F^2\le
8\pi_{\min}^{-2}(1+pp_{\max})\|D_F\|_F^2$ and
$\|\widehat P-P\|_{1,1}\le p\|\widehat P-P\|_F$.
\end{lemma}
\begin{proof}
The exact row identity is
\[
Q_{i,:}-P_{i,:}
=\frac{D_{F,i,:}-(D_F\mathbf1)_i P_{i,:}}{\widehat\pi_i}.
\]
Euclidean projection onto a closed convex set containing $P_{i,:}$
cannot increase the distance to that point. Squaring the identity,
using $\|x+y\|_2^2\le2\|x\|_2^2+2\|y\|_2^2$, and summing
establish the first bound. Since
$\|P_{i,:}\|_2^2\le p_{\max}\sum_jP_{ij}=p_{\max}$ and
$\|D_F\mathbf1\|_2^2\le p\|D_F\|_F^2$, the remaining Frobenius
bounds follow. The entrywise-sum bound is Cauchy--Schwarz over $p^2$
entries. For the unprojected $Q$ one also has
$\|Q-P\|_{1,1}\le4\pi_{\min}^{-1}\|D_F\|_{1,1}$ directly
from the row identity; we do not assume that bound survives an
unspecified postprocessing step.
\end{proof}
A sufficient condition for the denominator event is
$\|D_F\mathbf1\|_\infty\le\pi_{\min}/2$.
When $pp_{\max}=O(1)$, the Frobenius conversion loses no additional
ambient factor beyond $\pi_{\min}^{-2}$.

\section{Structured Covariance Estimation under Expanded Dimensions}
\label{appendix:pca_application}

\subsection{Observation model and assumptions}
For each task $m$, observe independent mean-zero vectors
$X_1^{(m)},\ldots,X_{n_m}^{(m)}\in\mathbb R^{p_m}$ with covariance
$\Sigma^{(m)}=L^{(m)}+S^{(m)}\succeq0$. Source and target samples
are independent, and
\[
\widehat\Sigma^{(m)}=\frac1{n_m}\sum_{i=1}^{n_m}
X_i^{(m)}X_i^{(m)\top}=\Sigma^{(m)}+W^{(m)}.
\]
For the statistical guarantees, assume the structured transfer and
source-anchor conditions of Theorem~\ref{thm:main}, or their approximate
version in Theorem~\ref{thm:main-robust}. The target statistical
estimator is a feasible approximately optimal fit to
$Y^{(2)}=\widehat\Sigma^{(2)}$; the numerical experiments evaluate
Algorithm~\ref{alg:transfer-altproj} separately.

\begin{assumption}[Uniform sub-Gaussian covariance model]
\label{asp:covariance-subgaussian}
For a constant $K\ge1$, all linear functionals satisfy
\[
\|u^\top X_i^{(m)}\|_{\psi_2}
\le K\sqrt{u^\top\Sigma^{(m)}u}
\quad\text{for every }u\in\mathbb R^{p_m}.
\]
Here $\|Z\|_{\psi_2}=\inf\{t>0:\mathbb E\exp(Z^2/t^2)\le2\}$.
The covariance scale $\lambda_m=\|\Sigma^{(m)}\|_2$ is retained in
all bounds; it is not implicitly fixed to one.
\end{assumption}
This is a joint, covariance-relative sub-Gaussian condition, stronger
than a uniform bound on individual coordinates. It includes centered
Gaussian vectors. No restriction $n_m\ge p_m$ is imposed for the
full bounds below.

\subsection{Noise bounds and proof}
\begin{lemma}[Covariance noise and source-space compression]
\label{lem:covariance-noise}
Under Assumption~\ref{asp:covariance-subgaussian}, for one task with
parameters $(p,n,\lambda)$,
\[
\begin{aligned}
\mathbb E\|W\|_2^2&\lesssim_K
\lambda^2\left(\frac pn+\frac{p^2}{n^2}\right),\\
\mathbb E\|W\|_{\max}^2&\lesssim_K
\lambda^2\left(\frac{\log(e p)}n+
\frac{\log^2(e p)}{n^2}\right).
\end{aligned}
\]
For orthonormal $p\times r_1$ frames $U_a,V_a$ independent of these
samples,
\[
\mathbb E[\|U_a^\top WV_a\|_2^2\mid U_a,V_a]
\lesssim_K\lambda^2\left(\frac{r_1}n+
\frac{r_1^2}{n^2}\right).
\]
\end{lemma}
\begin{proof}
We give the argument to record the dependence on dimension, covariance
scale, and the second Bernstein term; see also
\citet{vershynin2010introduction} for covariance concentration.
For any unit vector $u$, $(u^\top X_i)^2-
\mathbb E(u^\top X_i)^2$ has sub-exponential norm at most
$C K^2\lambda$. Scalar Bernstein and a $1/4$-net of the unit sphere
of cardinality at most $9^p$ give, for every $t\ge0$,
\[
\mathbb P\left\{\|W\|_2>
 C_K\lambda\left[\sqrt{\frac{p+t}{n}}+
                          \frac{p+t}{n}\right]\right\}
\le2e^{-t}.
\]
We used the symmetric-matrix net inequality
$\|W\|_2\le2\max_{u\in\mathcal N}|u^\top Wu|$.
Each entry $X_{i,j}X_{i,k}-\Sigma_{jk}$ also has
sub-exponential norm at most $CK^2\lambda$ by the product inequality
for sub-Gaussian variables. Bernstein and a union bound over $p^2$
entries give the same tail form with $p+t$ replaced by
$C\log(e p)+t$. Integrating the two tail bounds yields the stated
second moments, including the terms proportional to $n^{-2}$.

For the compressed matrix, condition on $U_a,V_a$ and let $H$ be an
orthonormal basis of their joint span, of dimension $k\le2r_1$.
Write $U_a=HC_U$, $V_a=HC_V$; the coefficient matrices have operator
norm one. The projected observations $H^\top X_i$ satisfy the same
sub-Gaussian condition with covariance $H^\top\Sigma H$ of norm at
most $\lambda$. Therefore
\[
\|U_a^\top WV_a\|_2\le\|H^\top WH\|_2,
\]
and the preceding covariance bound in dimension $k$ proves the last
claim. Independence from the target samples is essential for this
conditional argument.
\end{proof}

\subsection{Transfer and target-only error bounds}
\begin{corollary}[Covariance transfer rate]\label{cor:covariance-transfer}
Condition on source estimates satisfying the anchor conditions of
Theorem~\ref{thm:main}. Under Assumption~\ref{asp:covariance-subgaussian}
for the independent target observations, a feasible
$\varepsilon_{\mathrm{opt}}$-optimal target estimator obeys
\[
\begin{aligned}
&\mathbb E[\|\widehat L^{(2)}-L^{(2)}\|_F^2+
            \|\widehat S^{(2)}-S^{(2)}\|_F^2\mid\text{source}]\\
&\quad\lesssim_K\lambda_2^2\left\{
\frac{r_1^2}{n_2}+\frac{r_1^3}{n_2^2}
+\delta_{r,2}\left(\frac{p_2}{n_2}+\frac{p_2^2}{n_2^2}\right)
+\delta_{s,2}\left(\frac{\log(e p_2)}{n_2}+
                   \frac{\log^2(e p_2)}{n_2^2}\right)\right\}\\
&\qquad+\mathcal E_{\mathrm{src}}+\varepsilon_{\mathrm{opt}}.
\end{aligned}
\]
For approximate anchoring, add
$C(\epsilon_U^2+\epsilon_V^2)\|A^{(2)}\|_2^2$ under the
conditions of Theorem~\ref{thm:main-robust}.
\end{corollary}
\begin{proof}
Take conditional expectations in Theorem~\ref{thm:main} and apply
Lemma~\ref{lem:covariance-noise} to each of its three target-noise
terms. The source error is fixed under this conditioning.
The robust statement follows in exactly the same way from
Theorem~\ref{thm:main-robust}.
\end{proof}

For target-only constrained estimation under the corresponding
full-sparsity conditions, the same proof gives
\[
C_K\lambda_2^2\left\{
 r_2\left(\frac{p_2}{n_2}+\frac{p_2^2}{n_2^2}\right)
+s_2\left(\frac{\log(e p_2)}{n_2}+
          \frac{\log^2(e p_2)}{n_2^2}\right)\right\}
+\varepsilon_{\mathrm{base}}.
\]
For $n_2\ge p_2$, the simpler leading comparison is
\[
\begin{aligned}
\text{transfer: }&\frac{\lambda_2^2}{n_2}
\{r_1^2+\delta_{r,2}p_2+\delta_{s,2}\log(e p_2)\},\\
\text{target only: }&\frac{\lambda_2^2}{n_2}
\{r_2p_2+s_2\log(e p_2)\},
\end{aligned}
\]
with source and optimization contributions retained separately.
When $n_2<p_2$, the full bounds, not this simplification, apply.
The source terms can be controlled via Corollary~\ref{thm:chai2025_(1)}
on events satisfying its signal and regularity conditions.

\paragraph{What the comparison establishes.}
The rates reduce the cost of estimating inherited structure, but they
do not imply consistency from $\delta_{r,2}\ll r_1$ and
$\delta_{s,2}\ll s_1$ alone. All displayed contributions must vanish
in a consistency regime. In particular, with a nonvanishing covariance
scale and freely re-estimated coefficients, $r_1\asymp n_2$ makes the
term $r_1^2/n_2$ nonvanishing. A nonvanishing upper bound for a baseline
is not a lower bound proving statistical impossibility. No such
impossibility claim or minimax optimality assertion is made here.
The numerical low-sample experiments assess practical performance;
they do not remove the assumptions or higher-order terms above.

\section{Empirical Application Set-up}\label{append:experiment_details}


In this part, we summarize the basic setup of our synthetic empirical application in Section~\ref{sec:empirical_synthetic}, including the key notation, its interpretation, and the values used in our experiments, as shown in Table~\ref{table:final_draw_hparams}.

\vspace{5pt}
\begin{table}[H]
\small
\centering
\caption{Key dimensions, ranks, sample sizes, and convergence parameters for the synthetic enlarged-dimension experiment.} \label{table:final_draw_hparams}
\setlength{\tabcolsep}{8pt}
\renewcommand{\arraystretch}{1.1}
\resizebox{0.8\textwidth}{!}{
\begin{tabular}{llc}
\toprule
\textbf{Notation} & \textbf{Meaning} & \textbf{Value} \\
\midrule
$p_1$ & Source task ambient dimension & $10$ \\
$r_1$ & Rank of the source low-rank component $L^{(1)}$ & $3$ \\
$n_1$ & Number of source samples used to form $\widehat{\Sigma}^{(1)}$ & $500$ \\
$p_2$ & Target task ambient dimension after enlargement & $50$ \\
$\delta_{r,2}$ & Rank increment from source to target (innovation rank) & $1$ \\
$r_2=r_1+\delta_{r,2}$ & Rank of the target low-rank component $L^{(2)}$ & $4$ \\
$n_2$ & Target sample sizes used to form $\widehat{\Sigma}^{(2)}$ & $\{30,50,80,100,120,150,200,250,300\}$ \\
$\varepsilon$ & Convergence tolerance (Algorithm~\ref{alg:transfer-altproj}) & $10^{-6}$ \\
$T_{\max}$ & Maximum AltProj iterations & $100$ \\
$\texttt{num\_trials}$ & Monte Carlo repetitions per $n_2$ (for averaging) & $50$ \\
\bottomrule
\end{tabular}
}
\end{table}

\paragraph{Practical hyperparameter selection.}
The rank increment and sparse-edit budget are numerical inputs. In the
main synthetic study the structural parameters are set using the
data-generating model; this isolates estimation from model selection.
Validation over a grid is a practical alternative when these values are
unknown. A spectrum-based rule requires adequate signal separation and
is not guaranteed to identify all cross-block innovation directions.
Larger budgets may increase variance and change the separation and
regularity conditions; no automatic constant-factor robustness to
arbitrary over-specification is asserted. The numerical procedure does
not enforce the incoherence constraint defining the statistical estimator.

\section{Controlled Rotation Robustness Study}\label{appendix:rotation}

This study evaluates the numerical procedure under controlled subspace
mismatch. It complements, but does not by itself verify, the sufficient
conditions of Theorem~\ref{thm:main-robust}.

\subsection{Setup}\label{appendix:rotation_setup}
The reported rotation study uses source sample size $n_1=200$ and
source rank $r_1=4$; these differ from the main covariance experiment.
We rotate $k\in\{1,2,3,4\}$ source directions by
$\theta=\rho\pi/2$ and examine $n_2\in\{50,100,150\}$.
The error is $\|\sin\Theta(\widehat U,U_{\mathrm{target}})\|_F$,
with equal-dimensional evaluation subspaces. For each $(n_2,k)$,
$\rho^\star$ is the first tested strength at which the reported
non-transfer error is strictly lower than the transfer error.
``$>1.00$'' means that no such crossing occurred on the tested grid
through $90^\circ$; it is not an extrapolation beyond that range.

\subsection{Results}\label{appendix:rotation_results}

\begin{table}[h]
\centering
\small
\setlength{\tabcolsep}{10pt}
\renewcommand{\arraystretch}{1.1}
\begin{tabular}{c c c c c}
\toprule
$n_2$ & rotated dirs $k$ & switch strength $\rho^\star$ & Transfer $\sin(\Theta)$ at $\rho^\star$ & Non-Transfer $\sin(\Theta)$ at $\rho^\star$ \\
\midrule
$50$  & $1$ & $> 1.00$ & --- & --- \\
$50$  & $2$ & $> 1.00$ & --- & --- \\
$50$  & $3$ & $> 1.00$ & --- & --- \\
$50$  & $4$ & $0.80$  & $2.022$ & $1.981$ \\
\midrule
$100$ & $1$ & $> 1.00$ & --- & --- \\
$100$ & $2$ & $0.50$  & $1.261$ & $1.098$ \\
$100$ & $3$ & $0.40$  & $1.257$ & $1.102$ \\
$100$ & $4$ & $0.40$  & $1.405$ & $1.102$ \\
\midrule
$150$ & $1$ & $0.50$  & $0.910$ & $0.845$ \\
$150$ & $2$ & $0.40$  & $0.991$ & $0.842$ \\
$150$ & $3$ & $0.30$  & $0.889$ & $0.842$ \\
$150$ & $4$ & $0.30$  & $1.010$ & $0.848$ \\
\bottomrule
\end{tabular}
\caption{Regime-switch table for the rotation robustness study. For each combination of target sample size $n_2$ and number of rotated source directions $k$, the table reports the smallest rotation strength $\rho^\star$ at which the non-transfer estimator first achieves lower subspace error than transfer. Entries ``$>1.00$'' indicate no switch was observed even at full $90^\circ$ rotation; corresponding error values are omitted. Source sample size $n_1 = 200$, source rank $r_1 = 4$.}
\label{tab:rotation_robustness}
\end{table}

For $n_2=50$, no crossing is reported for $k=1,2,3$ through
$90^\circ$; with four rotated directions the first crossing is
$72^\circ$. For $n_2=100$, no crossing is reported for one direction,
while the first crossings for two, three, and four directions are
$45^\circ,36^\circ,36^\circ$, respectively. With $n_2=150$ they
occur at $45^\circ,36^\circ,27^\circ,27^\circ$.
Thus the observed crossover occurs earlier with more rotated directions
and with more target data. A first-crossing table does not establish
monotonicity of the entire error curves, statistical significance at
each grid point, or permanent dominance beyond a crossing.

\subsection{Interpretation}\label{appendix:rotation_interpretation}
Principal angles are basis-invariant subspace quantities. The number of
rotated directions and their angular magnitudes describe different
aspects of mismatch. For $k$ orthogonal rotation planes and a common
angle $\theta$, the anchor-block discrepancy satisfies
$\|\sin\Theta\|_F^2=k\sin^2\theta$; additional target innovation
directions can affect source--target principal angles and must be
accounted for in the construction.
Accordingly, the experiment varies both the extent and magnitude of
drift; it does not show that angle is irrelevant.
If the fitted innovation budget is held fixed, rotations may create
misspecification rather than increase the model's fitted rank.
The table supports robustness of the practical procedure in the
reported low-target-sample settings, not a uniform no-negative-transfer
guarantee or a theorem about the empirical crossover location.

\section{NYC Taxi Data-Calibrated Experiments}  \label{appendix:real_data}

We calibrate transition models using publicly released NYC taxi trip records,\footnote{Available at \href{https://www.nyc.gov/site/tlc/about/tlc-trip-record-data.page}{https://www.nyc.gov/site/tlc/about/tlc-trip-record-data.page}.} provided to the NYC Taxi and Limousine Commission (TLC). Each record contains the pick-up and drop-off districts of a single trip, which we treat as a one-step Markov transition between predefined districts. We report three complementary studies: an approximation comparison (Section~\ref{appendix:real_data_lrps}) showing that the low-rank-plus-sparse decomposition fits this dataset substantially better than a pure low-rank model; a semi-synthetic comparison (Section~\ref{appendix:real_data_semisynth}) of our alternating-minimization estimator against a spectral baseline on a fixed state space; and a transfer experiment (Section~\ref{appendix:real_data_transfer}) under ambient state-space growth that evaluates the practical procedure of Section~\ref{sec:transition-matrix-estimation}, in which a chain learned on a subnetwork is transferred to the full network.

\subsection{Dataset and Preprocessing}\label{appendix:real_data_setup}

We gather all trips from January 2024, yielding a total of $2{,}964{,}624$ records. For the model-justification and semi-synthetic studies (Sections~\ref{appendix:real_data_lrps}--\ref{appendix:real_data_semisynth}), we exclude rarely visited districts (those appearing fewer than $100$ times), leaving $74$ districts and yielding an empirical transition matrix of dimension $p=74$. This empirical transition matrix serves as a fixed calibration target $P^\star$, not an observed population truth. For the transfer experiment (Section~\ref{appendix:real_data_transfer}), we use a less aggressive filter to retain a richer state space and obtain $p_2 = 255$ active districts; the source task is constructed by restricting to a randomly selected $p_1 = 200$ subnetwork (see Section~\ref{appendix:real_data_transfer} for details).

\subsection{Justifying the Low-Rank-Plus-Sparse Model}\label{appendix:real_data_lrps}

We first examine whether $P^\star$ is well approximated by a purely low-rank structure or whether the additional sparse component is necessary.

We compare two structured approximations. For the \emph{low-rank model} \citep{zhang2019spectral}, we compute the rank-$r$ truncated SVD of the empirical frequency matrix $F^\star$ to obtain $F_L$, then derive the transition estimate $P_L$ via row-wise normalization (with simplex projection where needed). For the \emph{low-rank-plus-sparse model}, we apply Algorithm~\ref{alg:transfer-altproj} (run as a single-task estimator with no anchoring) to $F^\star$ to obtain $F_{LS}$, and derive $P_{LS}$ analogously. We set $r = 10$ and sparsity budget $s = 50$. Approximation errors are reported in Table~\ref{tab:real-data-approximation}: the low-rank-plus-sparse model reduces the Frobenius approximation error of $F$ by roughly a factor of $9$ and the per-row $\ell_1$ error of $P$ by a factor of $5$, supporting the usefulness of additional sparse structure for this calibration matrix. Since the low-rank-plus-sparse model has more parameters, this approximation comparison alone does not establish exact structural validity or out-of-sample superiority.

\begin{table}[h]
	\centering
	\begin{tabular}{c|c||c|c}
		\hline
		$\|F^\star-F_{LS}\|_F$ & $0.0037$  &  $\|F^\star-F_{L}\|_F$ & $0.0349$\\ \hline
		$\|P^\star-P_{LS}\|_{1,1}/p$& $0.0075$ & $\|P^\star-P_{L}\|_{1,1}/p$ & $0.0363$\\ \hline
	\end{tabular}
	\caption{Approximation error of the low-rank-plus-sparse model ($LS$) and the low-rank model ($L$) on the NYC taxi transition matrix.}
	\label{tab:real-data-approximation}
\end{table}

\subsection{Semi-Synthetic Comparison: Alternating Minimization vs.\ Spectral}\label{appendix:real_data_semisynth}

\begin{figure}[h]
\begin{center}
    \begin{tabular}{cc}
    \includegraphics[width=0.4\linewidth]{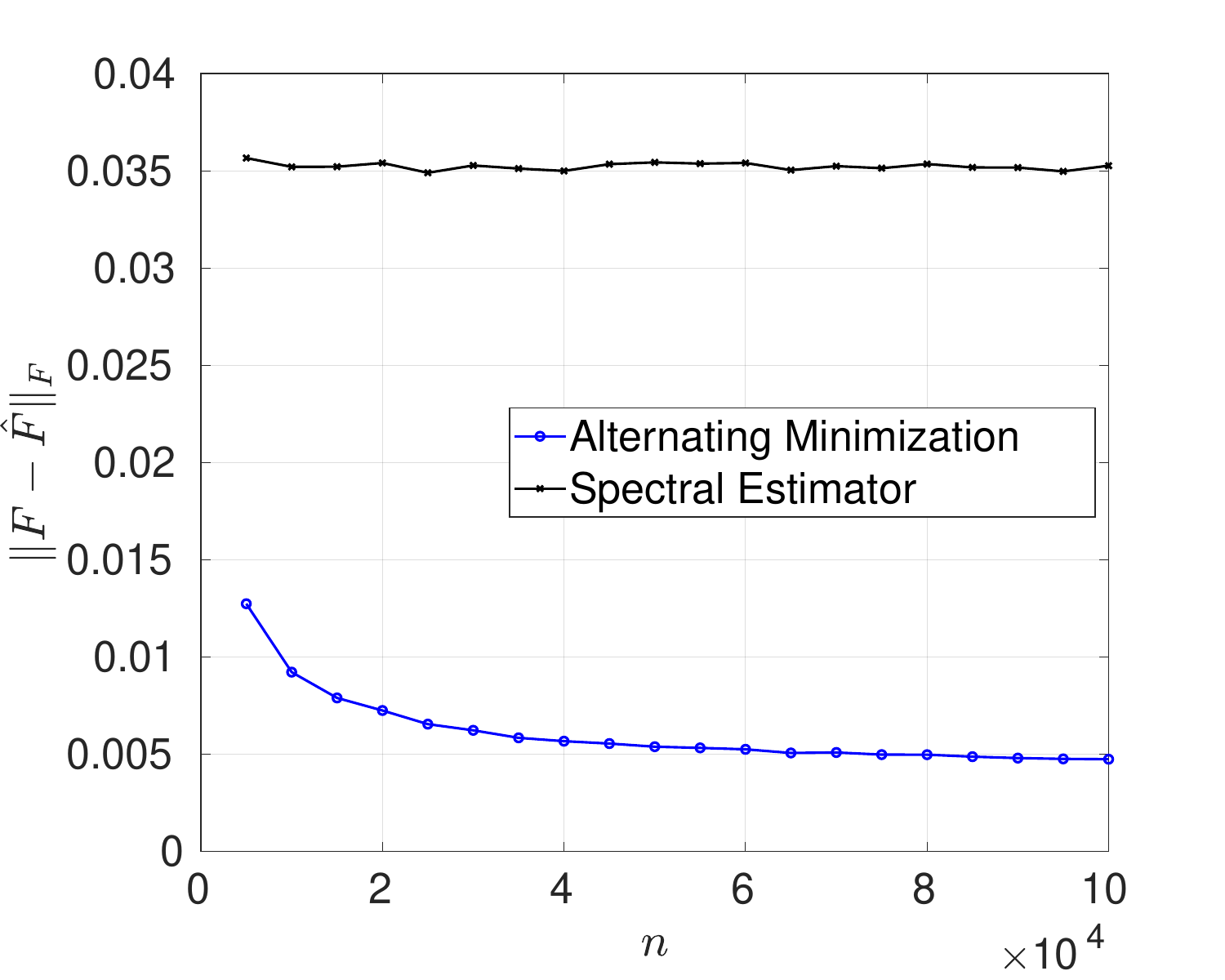}  &
    \includegraphics[width=0.4\linewidth]{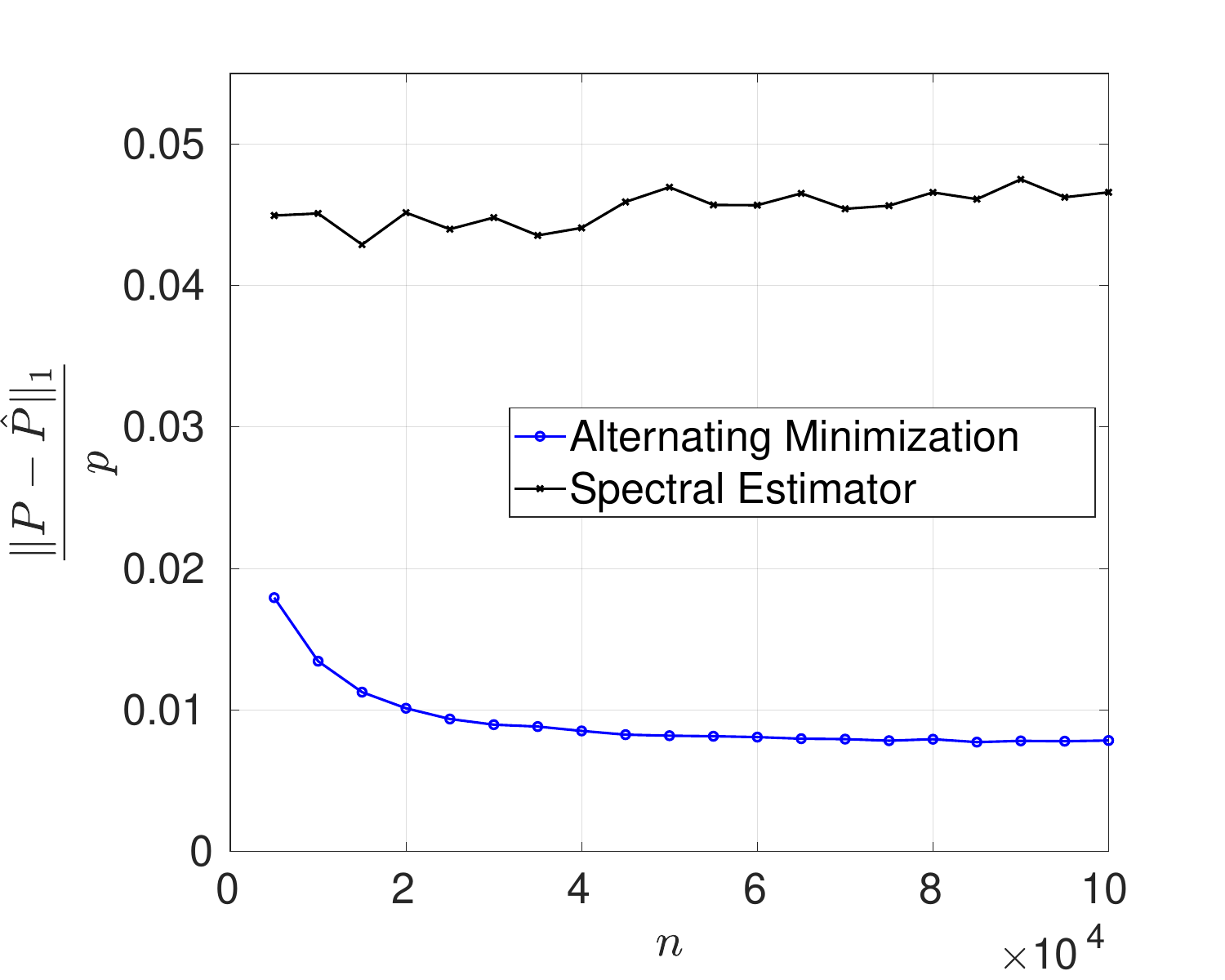} \\
    $\|F-\widehat F\|_F$ vs.\ $n$ & $\|P-\widehat P\|_{1,1}/p$ vs.\ $n$
    \end{tabular}
\end{center}
\caption{Spectral estimator (blue) vs.\ alternating minimization (black) on semi-synthetic NYC taxi data. Left: Frobenius error of the estimated frequency matrix. Right: per-row $\ell_1$ error of the estimated transition matrix. The alternating procedure improves over the tested range, while the spectral curve plateaus.}
\label{fig:real-data-comparison}
\end{figure}

Treating $P^\star$ as the ground truth, we generate Markov trajectories of length $n \in \{5000k : 1 \le k \le 20\}$, form the empirical frequency matrix, and apply two estimators: our alternating-minimization method (Algorithm~\ref{alg:transfer-altproj}, single-task version) and the spectral estimator of \citet{zhang2019spectral}. Results are averaged over $100$ trajectories per sample size and shown in Figure~\ref{fig:real-data-comparison}. The alternating-minimization estimator improves approximately monotonically with $n$ in both the Frobenius error of $F$ and the per-row $\ell_1$ error of $P$, while the spectral estimator exhibits a plateau over the tested range, consistent with approximation bias in its fixed-rank model. This interpretation is consistent with the approximation comparison in Table~\ref{tab:real-data-approximation}; the finite experiment does not independently prove an asymptotic bias limit.

\subsection{Transfer Experiment under Ambient State-Space Growth}\label{appendix:real_data_transfer}

We now test the central setting of Section~\ref{sec:transition-matrix-estimation}: transfer learning when the target chain has a strictly larger state space than the source. The aim is to measure gains of the practical solver on data-calibrated expansion, rather than certify its constrained optimality or independently observed cross-domain transfer.

\textbf{Source/target construction.}
We use the NYC taxi data with $p_2 = 255$ active districts, treating the corresponding empirical transition matrix as the target calibration matrix $P^{(2)\star}$. The source task is constructed by randomly selecting a subset of $p_1 = 200$ districts and restricting trips to those with both endpoints in the subset, producing a source calibration matrix $P^{(1)\star}$ on the smaller state space. The remaining $\delta_{p,2} = 55$ districts represent the genuinely new portion of the target state space, on which the source provides no information. The embedding $B(\cdot)$ in Section~\ref{sec:model} is realized by the natural inclusion of the source state indices into the target indexing.

\textbf{Sampling and estimation.}
For each trial, we draw a single source trajectory of length $n_1$ and a single target trajectory of length $n_2$, and form the empirical frequency matrices $\widehat F^{(1)}$ and $\widehat F^{(2)}$. The source anchors $(\widetilde U^{(1)}, \widetilde V^{(1)}, S_0)$ are obtained by running Algorithm~\ref{alg:transfer-altproj} (single-task version) on $\widehat F^{(1)}$, then embedding the resulting factors via $B(\cdot)$ as described in Section~\ref{sec:estimation}. We compare three estimators of $P^{(2)\star}$:
\begin{itemize}\itemsep1pt
\item \textbf{Transfer AltProj}: Algorithm~\ref{alg:transfer-altproj} applied to $\widehat F^{(2)}$ using anchors from the source, with rank increment $\delta_{r,2}$ and sparse-edit budget $\delta_{s,2}$;
\item \textbf{Non-transfer structured estimator}: Algorithm~\ref{alg:transfer-altproj} applied to $\widehat F^{(2)}$ with no anchoring, estimating the full target rank budget $r_2 = r_1 + \delta_{r,2}$ and a full-matrix sparsity budget from scratch;
\item \textbf{Truncated SVD}: rank-$r_2$ truncated SVD of $\widehat F^{(2)}$ followed by row-wise normalization.
\end{itemize}
Performance is measured by the principal-angle $\sin(\Theta)$ distance between the estimated and ground-truth left singular subspaces of $F^{(2)\star}$, which captures recovery of the dominant transition structure and is the natural metric for evaluating subspace-anchored transfer. Results are averaged over $20$ independent random source-subset selections and trajectory draws.

\textbf{Results.}
Table~\ref{tab:real-data-transfer} reports mean $\sin(\Theta)$ errors with $\pm 1$ standard deviations across the $20$ trials. Transfer AltProj achieves a mean error of $0.915 \pm 0.07$, compared to $1.679 \pm 0.09$ for the non-transfer structured estimator and $1.703 \pm 0.09$ for truncated SVD --- approximately a $45\%$ reduction relative to both baselines. The shared dataset supplies related calibration matrices, while the
random subset provides known coordinate correspondence without imposing
exact low-complexity inheritance. The experiment therefore supports
practical transfer under state-space expansion, not unrestricted
distribution shift. Source and target paths are simulated separately;
we do not treat a collection of individual taxi trips as one observed
stationary trajectory.

\begin{table}[h]
\centering
\begin{tabular}{l|c}
\hline
Estimator & Mean $\|\sin(\Theta)\|_F$ ($\pm$ 1 sd) \\ \hline
Transfer AltProj (ours) & $\mathbf{0.915 \pm 0.07}$ \\
Non-transfer structured estimator & $1.679 \pm 0.09$ \\
Truncated SVD & $1.703 \pm 0.09$ \\ \hline
\end{tabular}
\caption{Subspace recovery error on the NYC taxi transfer task ($p_1 = 200$, $p_2 = 255$), averaged over $20$ trials. Transfer AltProj reduces the principal-angle error by approximately $45\%$ relative to both non-transfer baselines.}
\label{tab:real-data-transfer}
\end{table}

\subsection{Metrics, computation, and data use}\label{appendix:reproducibility}
For equal-dimensional orthonormal frames $U,\widehat U\in\mathbb R^{p\times r}$,
our subspace metric is
\[
\|\sin\Theta(\widehat U,U)\|_F
=\left\{r-\|\widehat U^\top U\|_F^2\right\}^{1/2}
\in[0,\sqrt r].
\]
It measures aggregate subspace discrepancy, not an angle in degrees.
Matrix errors are Frobenius norms; per-row transition errors use
$\|P-\widehat P\|_{1,1}/p$. The NYC table reports means and one
standard deviation over 20 trials. The main covariance curves report
means over 50 trials; the exact figure assets determine whether
uncertainty bands are present.


\paragraph{Data provenance, permissions, and privacy.}
The NYC TLC is credited above as the provider of the public trip-record
release. This study reuses existing data and does not recruit
participants or collect new individual-level consent. Public release
should not be interpreted as a claim about each rider's consent.
The estimation targets aggregate pickup--dropoff district transitions,
not individual identification. Mobility records may nevertheless carry
privacy risks; reproduction and release should use only the information
needed for the aggregate analysis and follow the provider's terms.

\end{document}